\documentclass[10pt,journal,compsoc]{IEEEtran}
\ifCLASSOPTIONcompsoc
  \usepackage[nocompress]{cite}
\else
  \usepackage{cite}
\fi
\usepackage{adjustbox}
\usepackage{multirow}
\usepackage{subfigure}
\usepackage{epsfig,epstopdf}
\usepackage{bbding}
\usepackage{cite}
\usepackage[colorlinks]{hyperref}
\usepackage{graphicx}
\usepackage{amsmath}
\usepackage{amssymb}
\usepackage{balance}

\newcommand{\eg}{\textit{e}.\textit{g}.}
\newcommand{\yulun}[1]{\textcolor{black}{{#1}}}
\newcommand{\ylzhang}[1]{\textcolor{black}{{#1}}}

\ifCLASSINFOpdf
\else
\fi
\begin{document}

%
\title{Residual Dense Network for Image Restoration}
%
%
%
%

\author{Yulun Zhang,
        Yapeng Tian,
        Yu Kong,
        Bineng Zhong,
        and~Yun~Fu,~\IEEEmembership{Fellow,~IEEE}
\IEEEcompsocitemizethanks{\IEEEcompsocthanksitem Y. Zhang is with Department of ECE, Northeastern University, Boston, MA 02115. E-mail: yulun100@gmail.com
\IEEEcompsocthanksitem Y. Tian is with Department of Computer Science, University of Rochester, Rochester, NY 14627. E-Mail: yapengtian@rochester.edu
\IEEEcompsocthanksitem Y. Kong is with the B. Thomas Golisano College of Computing and Information Sciences, Rochester Institute of Technology, Rochester, NY 14623. E-Mail: yu.kong@rit.edu
\IEEEcompsocthanksitem B. Zhong is with School of Computer Science and Technology, Huaqiao University, Xiamen 361021, China. E-Mail: bnzhong@hqu.edu.cn
\IEEEcompsocthanksitem Y. Fu is with Department of ECE and College of CIS, Northeastern University, Boston, MA 02115. E-Mail: yunfu@ece.neu.edu}
} 

%
%

\markboth{Journal of \LaTeX\ Class Files,~Vol.~13, No.~9, September~2014}%
{Shell \MakeLowercase{\textit{et al.}}: Bare Demo of IEEEtran.cls for Computer Society Journals}
%



\IEEEtitleabstractindextext{%
\begin{abstract}
\yulun{Recently, deep convolutional neural network (CNN)} has achieved great success for image restoration (IR) and \yulun{provided} hierarchical features \yulun{at the same time}. However, most deep CNN based IR models do not make full use of the hierarchical features from the original low-quality images, thereby \yulun{resulting in relatively-low performance}. In this \yulun{work}, we propose a novel \yulun{and efficient} residual dense network (RDN) to address this problem in IR, by \ylzhang{making a better tradeoff between efficiency and effectiveness in exploiting the hierarchical features} from all the convolutional layers. Specifically, we propose residual dense block (RDB) to extract abundant local features via densely connected convolutional layers. RDB further allows direct connections from the state of preceding RDB to all the layers of current RDB, leading to a contiguous memory mechanism. To adaptively learn more effective features from preceding and current local features and stabilize the training of wider network, we proposed local feature fusion in RDB. After fully obtaining dense local features, we use global feature fusion to jointly and adaptively learn global hierarchical features in a holistic way. We demonstrate the effectiveness of RDN with \yulun{several} representative IR applications, single image super-resolution, Gaussian image denoising, image compression artifact reduction, \yulun{and image deblurring}. Experiments on benchmark \yulun{and real-world} datasets show that our RDN achieves favorable performance against state-of-the-art methods for each IR task \yulun{quantitatively and visually}.
\end{abstract}

}

\maketitle

\IEEEdisplaynontitleabstractindextext

%
\IEEEpeerreviewmaketitle

\IEEEraisesectionheading{\section{Introduction}\label{sec:introduction}}
\IEEEPARstart{S}{ingle} image restoration (IR) aims to generate a visually pleasing high-quality (HQ) image from its degraded low-quality (LQ) measurement \yulun{(\eg, downsaled, noisy, compressed, or/and blurred images). Image restoration plays an important and fundamental role and has been widely used in computer vision, ranging from security and surveillance imaging~\cite{zou2012very}, medical imaging~\cite{shi2013cardiac}, to image generation~\cite{karras2018progressive}. However, IR is very challenging, because the image degradation process is irreversible, resulting in an ill-posed inverse procedure. To tackle this problem, lots of works have been proposed, such as model-based~\cite{elad2006image,dong2012nonlocally,dong2015image,dong2019denoising} and learning-based~\cite{zhang2017learning,tai2017memnet} methods. Recently, deep convolutional neural network (CNN) has been widely investigated and achieves promising performance in various image restoration tasks, such as image super-resolution (SR)~\cite{zhang2006edge,zhang2012single,timofte2013anchored,timofte2014a+,peleg2014statistical,dong2014learning,schulter2015fast,huang2015single,kim2016accurate,tong2017image,zhang2018learning}, image denoising (DN)~\cite{chen2017trainable,mao2016image,zhang2017beyond,tai2017memnet,zhang2017learning,zhang2017ffdnet}, image compression reduction (CAR)~\cite{dong2015compression,chen2017trainable,zhang2017beyond}, and image deblurring~\cite{dong2012nonlocally,dong2015image,zhang2017learning,dong2019denoising}.}

\begin{figure}[t]
\centering
\centerline{
\subfigure[{ Residual block}]{
\label{fig:blocks_RB}
\includegraphics[scale = 0.68]{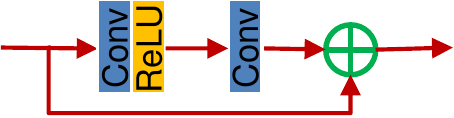}}
\subfigure[{ Dense block}]{
\label{fig:blocks_DB}
\includegraphics[scale = 0.68]{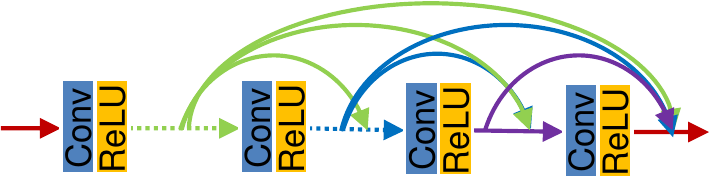}}
}
\centerline{
\subfigure[{ Residual dense block}]{
\label{fig:blocks_RDB}
\includegraphics[scale = 0.68]{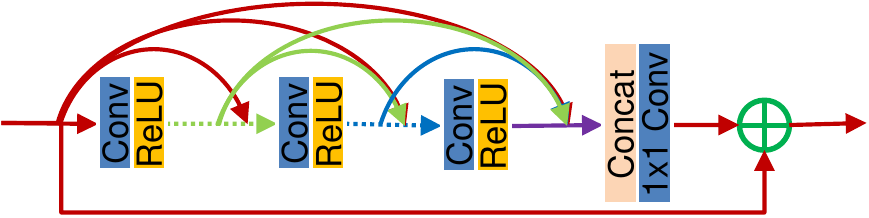}}
}

\vspace{-2mm}
\caption{Comparison of prior network structures (a,b) and our residual dense block (c). (a) Residual block in MDSR~\cite{lim2017enhanced}. (b) Dense block in SRDenseNet~\cite{tong2017image}. (c) Our \yulun{proposed residual dense block (RDB), which not only enables previous RDB to connect with each layer of current RDB, but also makes better use of local features.}} 
\label{fig:blocks_RB_DB_RDB}
\end{figure}

\yulun{The first challenge is how to introduce deep CNN for image restoration. Dong~et al.~\cite{dong2014learning}, for the first time, proposed SRCNN for image SR with three convolutional (Conv) layers and achieved significant improvement over previous methods. After firstly introducing CNN for image SR, Dong~et al.~\cite{dong2015compression} further applied CNN for other image restoration tasks, like image CAR. But, it's hard to train by stacking more Conv layers in SRCNN. To ease the difficulty of training deep network, Kim et al. proposed VDSR~\cite{kim2016accurate} and DRCN~\cite{kim2016deeply} by using gradient clipping, residual learning, or recursive-supervision. With newly investigated effective building modules and training strategies, better image SR performance could be further achieved. Incorporating residual learning, Zhang et al.~\cite{zhang2017beyond} proposed efficient denoising network. Memory block was proposed by Tai et al. to build MemNet for image restoration~\cite{tai2017memnet}. Later, Lim et al. proposed simplified residual block (Figure~\ref{fig:blocks_RB} by removing batch normalization layers and came up with a very deep network MDSR~\cite{lim2017enhanced}. Lim et al. further investigated a very wide network EDSR~\cite{lim2017enhanced} by using residual scaling~\cite{szegedy2017inception} to stabilizing the training. Convolutional layers in different depth have different size of receptive fields, resulting in hierarchical features. However, these features are not fully used by previous methods. Although MemNet~\cite{tai2017memnet} involved a gate unit in memory block to control short-term memory, it failed to build direct connections between local Conv layers and their subsequent ones. As a result, memory block didn't make full use of the features from all the Conv layers within the block either.}

\yulun{What's more, same or similar objects would appear differently in the images due to different scales, angles of view, and aspect ratios. Hierarchical features can capture such characteristics, which would contribute to better reconstruction. However, most learning-based methods (\eg, VDSR~\cite{kim2016accurate}, LapSRN~\cite{lai2017deep}, IRCNN~\cite{zhang2017learning}, and EDSR~\cite{lim2017enhanced}) neglect to pursue more clues by \ylzhang{making better use of} hierarchical features. Although MemNet~\cite{tai2017memnet} takes features from several memory block, it failed to extract multi-level features from the original LQ image (\eg, the LR image). Let's take image SR as an example, MemNet would pre-process the original input by interpolating it to desired size. Such a step would not only introduce extra artifacts (\eg, blurring artifacts), but also increase the computation complexity quadratically. Later, dense block (Figure~\ref{fig:blocks_DB}) was introduced in SRDenseNet for image SR~\cite{tong2017image}. The growth rate is a key part in dense block. Based on the investigation in DenseNet~\cite{huang2017densely} and our experiments (see Section~\ref{subsec:study_DCG}), higher growth rate contributes to better performance, resulting in wider networks. However, training a wider network with dense blocks would become harder. As investigated in EDSR~\cite{lim2017enhanced}, increasing feature map number above a certain level would make the training more difficult and numerically unstable.}

\yulun{To tackle the issues and limitations above, we propose a simple yet efficient residual dense network (RDN) (Figure~\ref{fig:RDN}) by \ylzhang{extracting} the hierarchical features from the original LQ image. Our RDN is built on our proposed residual dense block (Figure~\ref{fig:blocks_RDB}). A straightforward way to obtain hierarchical feature is to directly extract the output of each Conv layer in the LQ space. But, it's impractical, especially for a very deep network. Instead, we propose residual dense block (RDB), which consists of dense connected layers and local feature fusion (LFF) with local residual learning (LRL). Furthermore, the Conv layers of current RDB have direct access to the previous RDB, resulting in a contiguous state pass. We name it as contiguous memory mechanism, which further passes on information that needs to be preserved~\cite{huang2017densely}. So, in each RDB, LFF concatenates the states of preceding and current RDBs and adaptively extracts local dense features. Moreover, LFF helps to stabilize the training of wider networks with high growth rate. After obtaining multi-level local dense features, global feature fusion (GFF) is conducted to adaptively preserve the hierarchical features in a global way. As shown in Figures~\ref{fig:RDN} and~\ref{fig:RDB}, the original LR input directly connects each layer, leading to an implicit deep supervision~\cite{lee2015deeply}.}

In summary, our main contributions are three-fold:
\begin{itemize}
\item We propose a unified framework residual dense network (RDN) for high-quality image restoration. The network makes full use of all the hierarchical features from the original LQ image.
\end{itemize}
\begin{itemize}
\item We propose residual dense block (RDB), which can not only read state from the preceding RDB via a contiguous memory (CM) mechanism, but also \ylzhang{better} utilize all the layers within it via local dense connections. The accumulated features are then adaptively preserved by local feature fusion (LFF).
\end{itemize}
\begin{itemize}
\item We propose global feature fusion to adaptively fuse hierarchical features from all RDBs in the LR space. With global residual learning, we combine the shallow features and deep features together, resulting in global dense features from the original LQ image. 
\end{itemize}

A preliminary version of this work has been presented as a conference version~\cite{zhang2018residual}. In the current work, we incorporate additional contents in significant ways:
\begin{itemize}
\item We investigate a flexible structure of RDN and apply it for different IR tasks. Such IR applications allow us to further investigate the potential breadth of RDN.
\end{itemize}
\begin{itemize}
\item We investigate more details and add considerable analyses to the initial version, such as block connection, network parameter number, and running time.   
\end{itemize}
\begin{itemize}
\item We extend RDN for Gaussian image denoising, compression artifact reduction, \yulun{and image deblurring}. Extensive experiments \yulun{on benchmark and real-world data} demonstrate that our RDN still outperforms existing approaches in these IR tasks.    
\end{itemize}

\section{Related Work}
\yulun{Existing IR methods mainly include model-based~\cite{elad2006image,dong2012nonlocally,dong2015image,dong2019denoising} and learning-based~\cite{zhang2017learning,tai2017memnet} methods. Among them,} deep learning (DL)-based methods have achieved dramatic advantages against conventional methods in computer vision~\cite{zhang2017image,zhang2018density,zhang2018densely,li2018tell,timofte2017ntire,haris2018deep,ancuti2018ntire,blau20182018,yu2018crafting,wang2018esrgan,zhang2018image,liu2018multi,plotz2018neural,liu2018non}. Here, we focus on several representative image restoration tasks, such as image super-resolution (SR), denoising (DN), compression artifact reduction (CAR), \yulun{and image deblurring.}  

\subsection{Image Super-Resolution}
\yulun{Deep learning was first investigated for image SR. The most challenging part is how to formulize the mapping between LR and HR images into deep neural network. After constructing the basic networks for image SR, there're still challenging problems to solve, such as how to train bery deep and wide network. We will show how these challenges are solved or alleviated.} 

\yulun{Dong et al.~\cite{dong2014learning} proposed SRCNN, applying CNN into image SR for the first time. Based on SRCNN, lots of improvements have been down to pursue better performance. By using residual learning to ease the training of deeper networks, VDSR~\cite{kim2016accurate} and IRCNN~\cite{zhang2017learning} could train deeper networks with more stacked Conv layers. DRCN~\cite{kim2016deeply} also achieved such a deep network by sharing network parameters and recursive learning, which was further employed in DRRN~\cite{tai2017image}. However, these methods would pre-process the original input by interpolating it to desired size. Such a step would not only introduce extra artifacts (\eg, blurring artifacts), but also increase the computation complexity quadratically. As a result, extracting features from the interpolated LR images would not be able to build direct mapping from the original LR to HR images.}

\yulun{To address the problem above, Dong et al.~\cite{dong2016accelerating} built direct mapping from the original LR image to the HR target by introducing a transposed Conv layer (also known as deconvolution layer). Such a transposed Conv layer was further replaced by an efficient sub-pixel convolution layer in ESPCN~\cite{shi2016real}. Due to its efficiency, sub-pixel Conv layer was adopted in SRResNet~\cite{ledig2017photo}, which took advantage of residual learning~\cite{he2016deep}. By extracting features from the original LR input and upscaling the final LR features with transposed or sub-pixel convolution layer, they could either be capable of real-time SR (\eg, FSRCNN and ESPCN), or be built to be very deep/wide (\eg, SRResNet and EDSR). However, all of these methods neglect to adequately utilize information from each Conv layer, but only upscale the high-level CNN features for the final reconstruction.}

\begin{figure*}[htbp]
\centering{
\subfigure[{RDN for image super-resolution (SR).}]{
\label{fig:RDNIR_SR}
\includegraphics[width = 160.5mm]{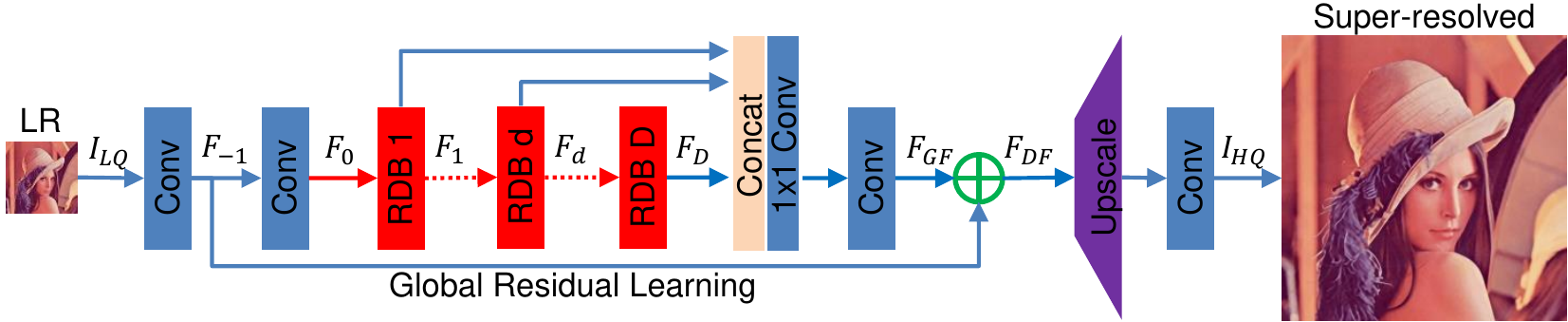}}
}
\centering{
\subfigure[{RDN for image denoising (DN), compression artifact reduction (CAR), and \yulun{deblurring. It should be noted that such a structure can also be used for image SR. But, it would consume more resources (\textit{e.g.}, GPU memory and running time).} Here, we use image DN as an example.}]{
\label{fig:RDNIR_DN}
\includegraphics[width = 160.5mm]{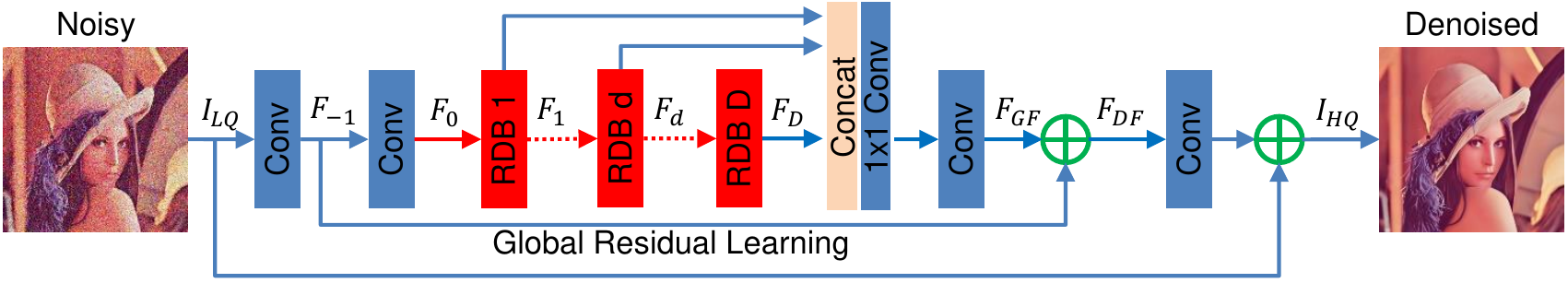}}
}

\vspace{-3mm}
\caption{The architecture of our proposed residual dense network (RDN) for image restoration.}
\label{fig:RDN} 
\vspace{-4mm}  
\end{figure*}

\subsection{Deep Convolutional Neural Network (CNN)}
LeCun et al. integrated constraints from task domain to enhance network generalization ability for handwritten zip code recognition~\cite{lecun1989backpropagation}, which can be viewed as the pioneering usage of CNNs. Later, various network structures were proposed with better performance, such as AlexNet~\cite{krizhevsky2012imagenet}, VGG~\cite{simonyan2014very}, and GoogleNet~\cite{szegedy2015going}. Recently, He et al.~\cite{he2016deep} investigated the powerful effectiveness of network depth and proposed deep residual learning for very deep trainable networks. Such a very deep residual network achieves significant improvements on several computer vision tasks, like image classification and object detection. Huang et al. proposed DenseNet, which allows direct connections between any two layers within the same dense block~\cite{huang2017densely}. With the local dense connections, each layer reads information from all the preceding layers within the same dense block. The dense connection was introduced among memory blocks~\cite{tai2017memnet} and dense blocks~\cite{tong2017image}. More differences between DenseNet/SRDenseNet/MemNet and our RDN would be discussed in Section~\ref{sec:differences}.

\subsection{Deep Learning for Image Restoration}
\yulun{Generally, there are two categories of methods for image restoration: model-based~\cite{elad2006image,dong2012nonlocally,dong2015image,dong2019denoising} and learning-based~\cite{zhang2017learning,tai2017memnet} methods. The model-based methods often formulate the image restoration problems as optimization ones. Numerous regularizers have been investigated to search for better solutions, such as sparsity-based regularizers with learned dictionaries~\cite{elad2006image} and nonlocal self-similarity inspired ones~\cite{dong2015image}. Also, instead of using explicitly designed regularizers, Dong et al.~\cite{dong2019denoising} proposed a denoising prior driven network for image restoration. Then, we give more details about learning-based methods.}

Dong et al.~\cite{dong2015compression} proposed ARCNN for image compression artifact reduction (CAR) with several stacked convolutional layers. Mao et al.~\cite{mao2016image} proposed residual encoder-decoder networks (RED) with symmetric skip connections, which made the network go deeper (up to 30 layers). Zhang et al.~\cite{zhang2017beyond} proposed DnCNN to learn mappings from noisy images to noise and further improved performance by utilizing batch normalization~\cite{ioffe2015batch}. Zhang et al.~\cite{zhang2017learning} proposed to learn deep CNN denoiser prior for image restoration (IRCNN) by integrating CNN denoisers into model-based optimization method. However, such methods have limited network depth (\eg, 30 for RED, 20 for DnCNN, and 7 for IRCNN), limiting the network ability. Simply stacking more layers cannot reach better results due to gradient vanishing problem. On the other hand, by using short term and \yulun{long-term} memory, Tai et al.~\cite{tai2017memnet} proposed MemNet for image restoration, where the network depth reached 212 but obtained limited improvement over results with 80 layers. For 31$\times$31 input patches from 91 images, training an 80-layer MemNet takes 5 days using 1 Tesla P40 GPU~\cite{tai2017memnet}. 

The aforementioned DL-based image restoration methods have achieved significant improvement over conventional methods, but most of them lose some useful hierarchical features from the original LQ image. Hierarchical features produced by a very deep network are useful for image restoration tasks (\eg, image SR). To fix this case, we propose residual dense network (RDN) to extract and adaptively fuse features from all the layers in the LQ space efficiently.

\section{Residual Dense Network for IR}
In Figure~\ref{fig:RDN}, we show our proposed RDN for image restoration, including image super-resolution (see Figure~\ref{fig:RDNIR_SR}), denoising, compression artifact reduction, \yulun{and deblurring} (see Figure~\ref{fig:RDNIR_DN}). 
\subsection{Network Structure}
\label{sec:network}
we mainly take image SR as an example and give specific illustrations for image DN and CAR cases.

\textbf{RDN for image SR}. As shown in Figure~\ref{fig:RDNIR_SR}, our RDN mainly consists of four parts: shallow feature extraction net, residual dense blocks (RDBs), dense feature fusion (DFF), and finally the up-sampling net (UPNet). Let's denote $I_{LQ}$ and $I_{HQ}$ as the input and output of RDN. Specifically, we use two Conv layers to extract shallow features. The first Conv layer extracts features $F_{-1}$ from the LQ input.
\begin{align}
\begin{split}
\label{eq:SFE1}
F_{-1}=H_{SFE1}\left ( I_{LQ} \right ),
\end{split}
\end{align}
where $H_{SFE1}\left ( \cdot  \right )$ denotes convolution operation. $F_{-1}$ is then used for further shallow feature extraction and global residual learning. So, we can further have
\begin{align}
\begin{split}
\label{eq:SFE2}
F_{0}=H_{SFE2}\left ( F_{-1} \right ),
\end{split}
\end{align}
where $H_{SFE2}\left ( \cdot  \right )$ denotes convolution operation of the second shallow feature extraction layer and is used as input to residual dense blocks. Supposing we have $D$ residual dense blocks, the output $F_{d}$ of the $d$-th RDB can be obtained by
\begin{align}
\begin{split}
\label{eq:F_d_RDN}
F_{d}&=H_{RDB,d}\left ( F_{d-1} \right )\\
&=H_{RDB,d}\left ( H_{RDB,{d-1}}\left ( \cdots \left ( H_{RDB,1}\left ( F_{0} \right ) \right ) \cdots \right ) \right ),
\end{split}
\end{align}
where $H_{RDB,d}$ denotes the operations of the $d$-th RDB. $H_{RDB,d}$ can be a composite function of operations, such as convolution and rectified linear units (ReLU)~\cite{glorot2011deep}. As $F_{d}$ is produced by the $d$-th RDB fully utilizing each convolutional layers within the block, we can view $F_{d}$ as local feature. More details about RDB will be given in Section~\ref{subsec:RDB}.

After extracting hierarchical features with a set of RDBs, we further conduct dense feature fusion (DFF), which includes global feature fusion (GFF) and \yulun{global residual learning}. DFF makes full use of features from all the preceding layers and can be represented as
\begin{align}
\begin{split}
\label{eq:DFF}
F_{DF}=H_{DFF}\left ( F_{-1},F_{0},F_{1},\cdots ,F_{D} \right ),
\end{split}
\end{align}  
where $F_{DF}$ is the output feature-maps of DFF by utilizing a composite function $H_{DFF}$. More details about DFF will be shown in Section~\ref{subsec:DFF}. 

After extracting local and global features in the LQ space, we stack an up-sampling net (UPNet) in the HQ space. Inspired by~\cite{lim2017enhanced}, we utilize ESPCN~\cite{shi2016real} in UPNet followed by one Conv layer. The output of RDN can be obtained by
\begin{align}
\begin{split}
\label{eq:I_SR}
I_{HQ}=H_{RDN}\left ( I_{LQ} \right ),
\end{split}
\end{align}  
where $H_{RDN}$ denotes the function of our RDN.

\textbf{RDN for image DN, CAR, \yulun{and Deblurring}}. When we apply our RDN to image DN, CAR, \yulun{and Deblurring}, the resolution of the input and output keep the same. As shown in Figure~\ref{fig:RDNIR_DN}, we remove the upscaling module in UPNet and obtain the final HQ output via residual learning 
\begin{align}
\begin{split}
\label{eq:I_DN}
I_{HQ}=H_{RDN}\left ( I_{LQ} \right ) + I_{LQ}.
\end{split}
\end{align}

\yulun{Unlike image SR, here, we use residual learning connecting input and output for faster training. It should be noted that the network structure in Figure~\ref{fig:RDNIR_DN} is also suitable for image SR. But, it would consume more GPU memory and running time. As a result, we choose the network structure in Figure~\ref{fig:RDNIR_SR} for image SR.}

\begin{figure}[tpb]
\centering
\includegraphics[scale = 0.8]{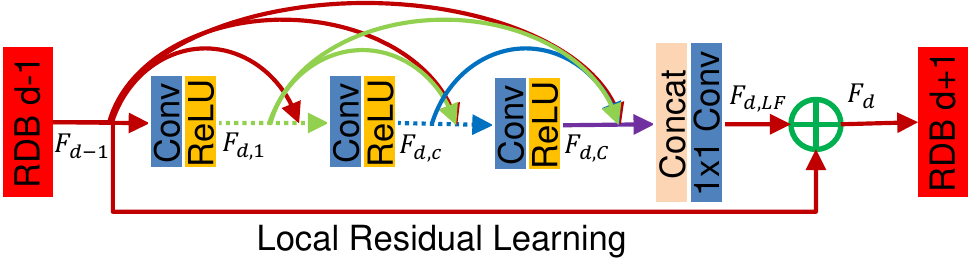}
\vspace{-3mm}
\caption{Residual dense block (RDB) architecture. \yulun{We denote the connections between the ($d$-$1$)-th RDB and the following convolutional layers as contiguous memory (CM) mechanism}.}
\label{fig:RDB} 
\vspace{-4mm} 
\end{figure}

\subsection{Residual Dense Block}
\label{subsec:RDB}
   
We present details about our proposed residual dense block (RDB) shown in Figure~\ref{fig:RDB}. Our RDB contains dense connected layers, local feature fusion (LFF), and local residual learning, leading to a contiguous memory (CM) mechanism.     

\textbf{Contiguous memory (CM) mechanism}. The idea behind CM is that we \ylzhang{aim at fusing} information from all the Conv layers as much as possible. But, directly fuse feature maps from all the Covn layers is not practical, because it would stack huge amount of features. Instead, we turn to first adaptively fuse information locally and then pass them to the following feature fusions. It is realized by passing the state of preceding RDB to each layer of the current RDB. Let $F_{d-1}$ and $F_{d}$ be the input and output of the $d$-th RDB respectively and both have G$_{0}$ feature-maps. The output of $c$-th Conv layer of $d$-th RDB can be formulated as
\begin{align}
\begin{split}
\label{eq:F_d_c}
F_{d,c}=\sigma \left ( W_{d,c}\left [ F_{d-1},F_{d,1},\cdots ,F_{d,c-1} \right ]  \right ),
\end{split}
\end{align}
where $\sigma$ denotes the ReLU~\cite{glorot2011deep} activation function. $W_{d,c}$ is the weights of the $c$-th Conv layer, where the bias term is omitted for simplicity. We assume $F_{d,c}$ consists of G (also known as growth rate~\cite{huang2017densely}) feature-maps. $\left [ F_{d-1},F_{d,1},\cdots ,F_{d,c-1} \right ]$ refers to the concatenation of the feature-maps produced by the $(d-1)$-th RDB, convolutional layers $1,\cdots ,\left ( c-1 \right )$ in the $d$-th RDB, resulting in G$_{0}$+$\left ( c-1 \right )\times $G feature-maps. The outputs of the preceding RDB and each layer have direct connections to all subsequent layers, which not only preserves the feed-forward nature, but also extracts local dense feature.

\textbf{Local feature fusion (LFF)}. We apply LFF to adaptively fuse the states from preceding RDB and the whole Conv layers in current RDB. As analyzed above, the feature-maps of the $\left ( d-1 \right )$-th RDB are introduced directly to the $d$-th RDB in a concatenation way, it is essential to reduce the feature number. On the other hand, inspired by~MemNet~\cite{tai2017memnet}, we introduce a $1\times 1$ convolutional layer to adaptively control the output information. We name this operation as local feature fusion (LFF) formulated as
\begin{align}
\begin{split}
\label{eq:F_d_LF}
F_{d,LF}=H_{LFF}^{d}\left ( \left [ F_{d-1},F_{d,1},\cdots ,F_{d,c},\cdots ,F_{d,C}\right ] \right ),
\end{split}
\end{align}
where $H_{LFF}^{d}$ denotes the function of the $1\times 1$ Conv layer in the $d$-th RDB. We also find that as the growth rate G becomes larger, very deep dense network without LFF would be hard to train. However, larger growth rate further contributes to the performance, which will be detailed in Section~\ref{subsec:study_DCG}.

\textbf{Local residual learning (LRL)}. We introduce LRL in RDB to further improve the information flow and allow larger growth rate, as there are several convolutional layers in one RDB. The final output of the $d$-th RDB can be obtained by
\begin{align}
\begin{split}
\label{eq:F_d_RDB}
F_{d}=F_{d-1}+F_{d,LF}.
\end{split}
\end{align}
It should be noted that LRL can also further improve the network representation ability, resulting in better performance. We introduce more results about LRL in Section~\ref{subsec:ablation}. Because of the dense connectivity and local residual learning, we refer to this block architecture as residual dense block (RDB). More differences between RDB and original dense block~\cite{huang2017densely} would be summarized in Section~\ref{sec:differences}.

\subsection{Dense Feature Fusion}
\label{subsec:DFF}
After extracting local dense features with a set of RDBs, we further propose dense feature fusion (DFF) to exploit hierarchical features in a global way. DFF consists of global feature fusion (GFF) and \yulun{global residual learning}. 

\textbf{Global feature fusion (GFF)}. We propose GFF to extract the global feature $F_{GF}$ by fusing features from all the RDBs
\begin{align}
\begin{split}
\label{eq:GFF}
F_{GF}=H_{GFF}\left ( \left [ F_{1},\cdots ,F_{D} \right ] \right ),
\end{split}
\end{align}
where $\left [ F_{1},\cdots ,F_{D} \right ]$ refers to the concatenation of feature maps produced by residual dense blocks $1,\cdots ,D$. $H_{GFF}$ is a composite function of $1\times 1$ and $3\times 3$ convolution. The $1\times 1$ convolutional layer is used to adaptively fuse a range of features with different levels. The following $3\times 3$ convolutional layer is introduced to further extract features for global residual learning, which has been demonstrated to be effective in~\cite{ledig2017photo}.

\textbf{Global residual learning}. We then utilize \yulun{global residual learning} to obtain the feature-maps before conducting up-scaling by 
\begin{align}
\begin{split}
\label{eq:DFF_GF}
F_{DF}=F_{-1}+F_{GF},
\end{split}
\end{align}
where $F_{-1}$ denotes the shallow feature-maps. All the other layers before global feature fusion are \ylzhang{extensively} utilized with our proposed residual dense blocks (RDBs). RDBs produce multi-level local dense features, which are further adaptively fused to form $F_{GF}$. After global residual learning, we obtain deep dense feature $F_{DF}$.

It should be noted that Tai~et al.~\cite{tai2017memnet} utilized long-term dense connections in MemNet to recover more high-frequency information. However, in the memory block~\cite{tai2017memnet}, the preceding layers don't have direct access to all the subsequent layers. The local feature information is not fully used, limiting the ability of long-term connections. In addition, MemNet extracts features in the HQ space, increasing computational complexity. While, inspired by~\cite{dong2016accelerating,shi2016real,lai2017deep,lim2017enhanced}, we extract local and global features in the LQ space. More differences between our proposed RDN and MemNet would be shown in Section~\ref{sec:differences}. We would also demonstrate the effectiveness of global feature fusion in Section~\ref{subsec:ablation}.

\subsection{Implementation Details}
\label{sec:implement}
In our proposed RDN, we set $3\times 3$ as the size of all convolutional layers except that in local and global feature fusion, whose kernel size is $1\times 1$. For convolutional layer with kernel size $3\times 3$, we pad zeros to each side of the input to keep size fixed. Shallow feature extraction layers, local and global feature fusion layers have G$_{0}$=64 filters. Other layers in each RDB has G=64 filters and are followed by ReLU~\cite{glorot2011deep}. For image SR, following~\cite{lim2017enhanced}, we use ESPCNN~\cite{shi2016real} to upscale the coarse resolution features to fine ones for the UPNet. For image DN and CAR, the up-scaling module is removed from UPNet. The final Conv layer has $3$ output channels, as we output color HQ images. However, the network can also process gray images, for example, when we apply RDN for gray-scale image denoising.

\begin{figure*}[htbp]
\scriptsize
\centering
\centerline{
\subfigure[]{
\label{fig:investeD}
\includegraphics[width = 59.5mm]{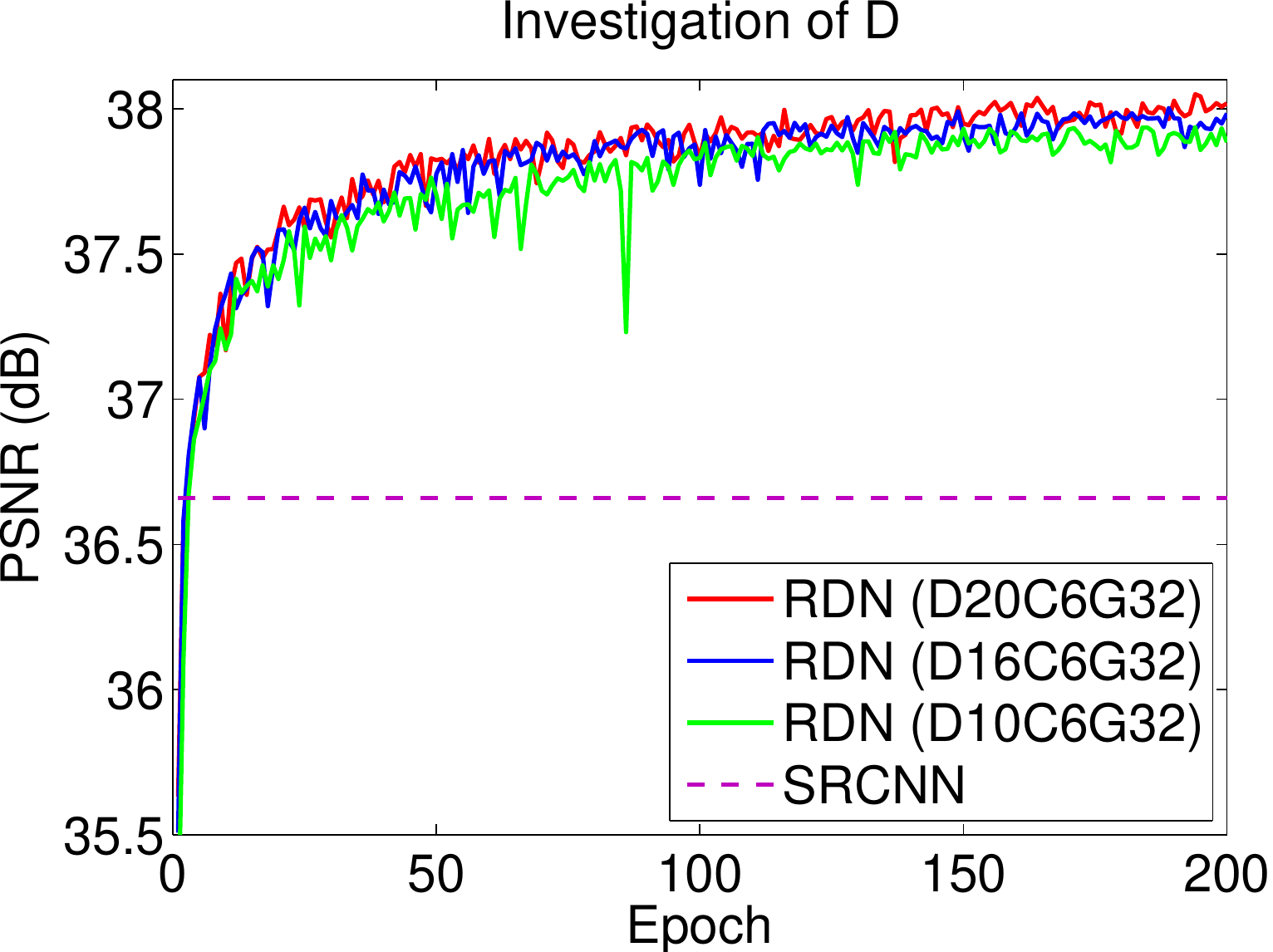}}
\subfigure[]{
\label{fig:investeC}
\includegraphics[width = 59.5mm]{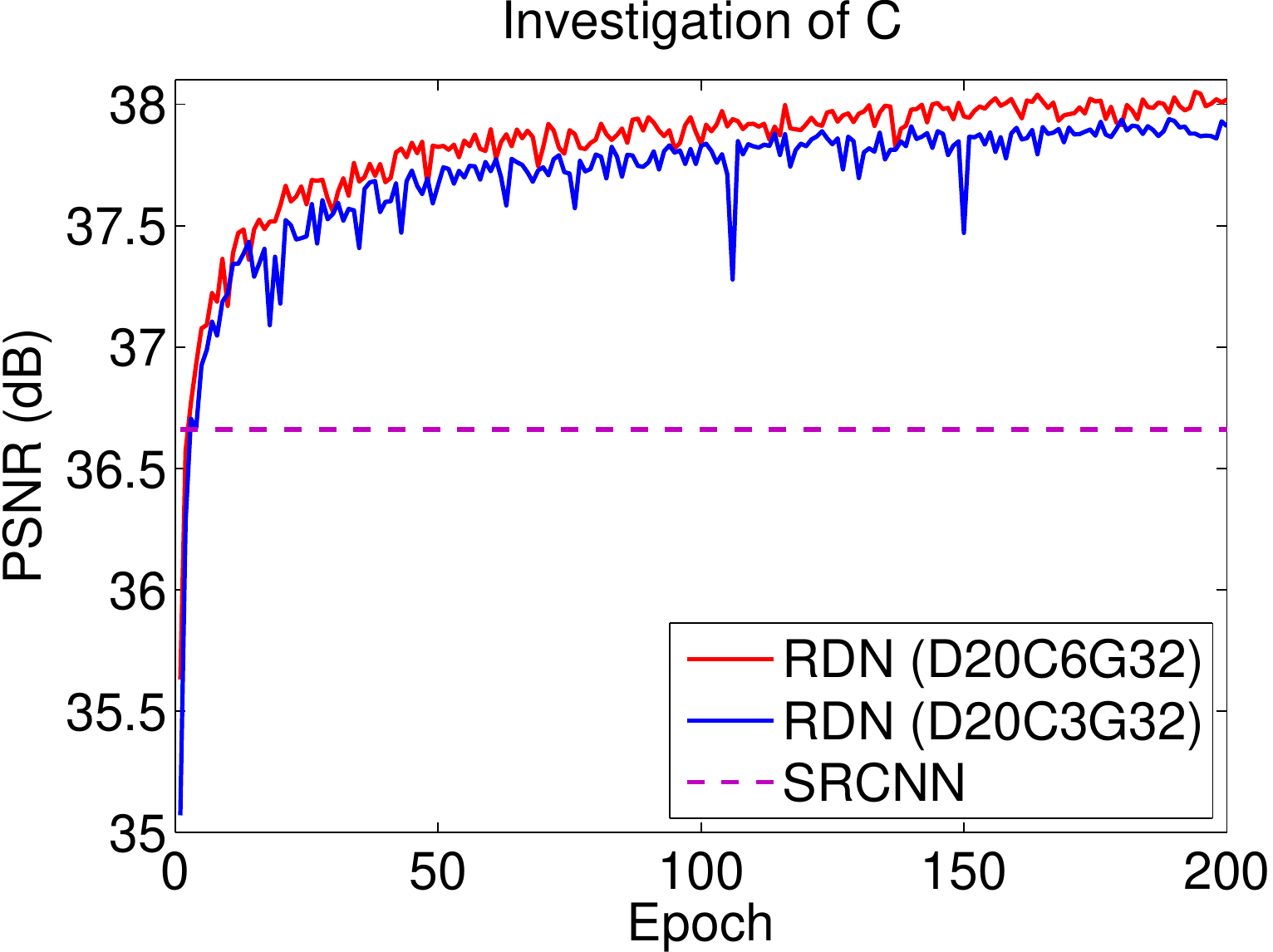}}
\subfigure[]{
\label{fig:investeG}
\includegraphics[width = 59.5mm]{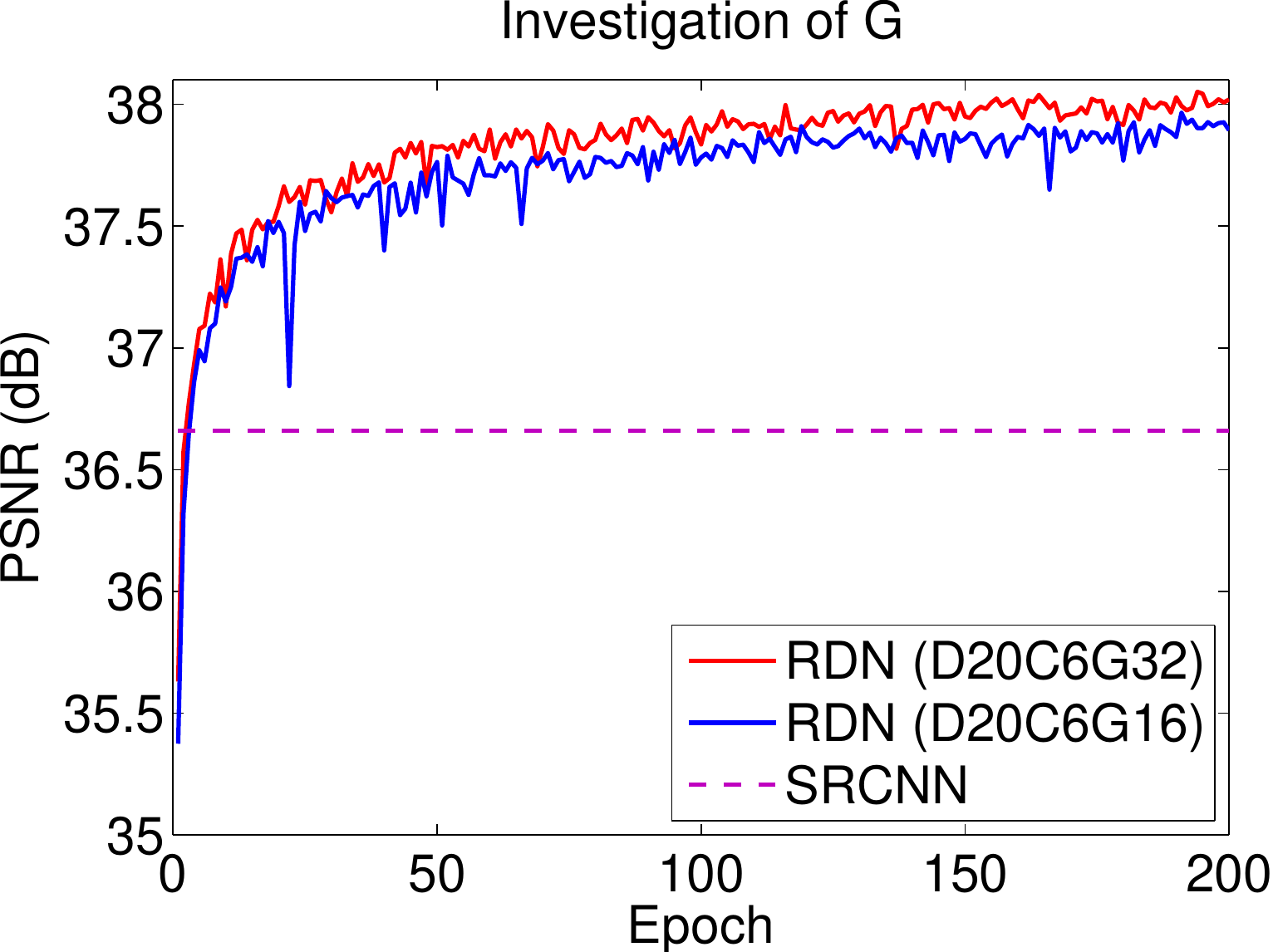}}
}
\vspace{-3mm}
\caption{Convergence analysis of RDN for image SR ($\times2$) with different values of D, C, and G.}  
\label{fig:study_D_C_G}
\end{figure*}

\section{Differences with Prior Works}
\label{sec:differences}
Here, we give more details about the differences between our RDN and several related representative works. \yulun{We further demonstrate those differences make our RDN more effective in Sections~\ref{sec:network_investigation} and \ref{sec:results}.}

\textbf{Difference to DenseNet}. \yulun{First, DenseNet~\cite{huang2017densely} is widely used in high-level computer vision tasks (\eg, image classification). While RDN is designed more specific for image restoration without using some operations in DenseNet, such as batch normalization and max pooling. Second, DenseNet places transition layers into two adjacent dense blocks. In RDN, the Conv layers are connected by local feature fusion (LFF) and local residual learning. Consequently, the $(d-1)$-th RDB has direct access to each layer in the $d$-th RDB and also contributes to the input of $(d+1)$-th RDB. Third, we adopt GFF to \ylzhang{make better use of} hierarchical features, which are neglected in DenseNet.}

\textbf{Difference to SRDenseNet}. \yulun{First, SRDenseNet~\cite{tong2017image} introduces the basic dense block with batch normalization from DenseNet~\cite{huang2017densely}. Our residual dense block (RDB) improves it in several ways: (1). We propose contiguous memory (CM) mechanism, building direct connections between the preceding RDB and each layer of the current RDB. (2). Our RDB can be easily extended to be wider and be easy to train with usage of local feature fusion (LFF). (3). RDB utilizes local residual learning (LRL) to further encourage the information flow. Second, SRDenseNet densely connects dense blocks, while there are no dense connections among RDBs. We use global feature fusion (GFF) and global residual learning to extract hierarchical features, based on the fully extracted extracted local features by RDBs. Third, SRDenseNet aims to minimize $L_2$ loss. However, bing more powerful for performance and convergence~\cite{lim2017enhanced}, $L_1$ loss function is utilized in RDN for each image restoration task.}

\textbf{Difference to MemNet}. \yulun{In addition to the different choices for loss fuction between MemNet~\cite{tai2017memnet} and RDN, we mainly summarize additional three differences. First, MemNet has to interpolate the original LR to the desired size, introducing extra blurry artifacts. While, RDN directly extracts features from the original LR input, which helps to reduce computational complexity and contribute to better performance. Second, in MemNet, most Conv layers within one recursive unit have no direct access to their preceding layers or memory block. While, each Conv layer receives information from preceding RDB and outputs feature for the subsequent layers within same RDB. Plus, LRL further improves the information flow and performance. Third, in MemNet, the memory block neglects to fully utilize the features from preceding block and Conv layers within it. Although dense connections are adopted to connect memory blocks, MemNet fails to extract hierarchical features from the original LR images. However, with the local dense features extracted by RDBs, our RDN further fuses the hierarchical features from the whole preceding layers in a global way in the LR space.}

\begin{table}[tpb]
\scriptsize
\centering
\begin{center}
\caption{PSNR (dB) comparisons under different block connections. The results are obtained with Bicubic (\textbf{BI}) degradation model for image SR ($\times4$).}
\label{tab:results_dense_connections_block}
\vspace{-3mm}
\begin{tabular*}{83.5mm}{@{\extracolsep{-0.928mm}}|c|c|c|c|c|c|c|c|c|c|c|c|c|c|c|c|c|}
\hline
Block connection & \multicolumn{2}{c|}{Dense connections}  &  \multicolumn{2}{c|}{Contiguous memory}
\\
\hline
\multirow{2}{*}{Datasets} & SRDenseNet &  MemNet & RDN & RDN 
\\
& [\textcolor{green}{30}] & [\textcolor{green}{25}] & (w/o LRL) & (with LRL) 
\\
\hline
Set5~\cite{bevilacqua2012low}
& 32.02
 & 31.74
  & 32.54
   & 32.61
                                 
\\
\hline
Set14~\cite{zeyde2012single}
& 28.50
 & 28.26
  & 28.87
   & 28.93
                                 
\\
\hline
B100~\cite{martin2001database}
& 27.53
 & 27.40
  & 27.75
   & 27.80
                                 
\\
\hline
Urban100~\cite{huang2015single}
& 26.05
 & 25.50
  & 26.72
   & 26.85
                                 
\\
\hline
            
\end{tabular*}
\end{center}
\vspace{-4mm} 
\end{table}

\section{Network Investigations}
\label{sec:network_investigation}
\subsection{Study of D, C, and G.}
\label{subsec:study_DCG}
In this subsection, we investigate the basic network parameters: the number of RDB (denote as D for short), the number of Conv layers per RDB (denote as C for short), and the growth rate (denote as G for short). We use the performance of SRCNN~\cite{dong2016image} as a reference. As shown in Figures~\ref{fig:investeD} and~\ref{fig:investeC}, larger D or C would lead to higher performance. This is mainly because the network becomes deeper with larger D or C. As our proposed LFF allows larger G, we also observe larger G (see Figure~\ref{fig:investeG}) contributes to better performance. On the other hand, RND with smaller D, C, or G would suffer some performance drop in the training, but RDN would still outperform SRCNN~\cite{dong2016image}. More important, our RDN allows deeper and wider network, where more hierarchical features are extracted for higher performance.       

\begin{table}[htbp]
\scriptsize
\centering
\begin{center}
\caption{Ablation investigation of contiguous memory (CM), local residual learning (LRL), and global feature fusion (GFF). We observe the best performance (PSNR) on Set5 with scaling factor $\times2$ in 200 epochs.} 
\label{tab:results_ablation}
\vspace{-3mm}

\begin{tabular*}{82.9mm}{@{\extracolsep{-0.75mm}}|c|c|c|c|c|c|c|c|c|}
\hline
 & \multicolumn{8}{c|}{Different combinations of CM, LRL, and GFF} 
\\ 
\hline  
\hline
CM  & \XSolid & \Checkmark & \XSolid & \XSolid & \Checkmark & \Checkmark & \XSolid & \Checkmark
\\
LRL & \XSolid & \XSolid   & \Checkmark & \XSolid & \Checkmark & \XSolid & \Checkmark & \Checkmark
\\
GFF & \XSolid & \XSolid   & \XSolid & \Checkmark & \XSolid & \Checkmark & \Checkmark & \Checkmark
\\
\hline
\hline
PSNR & 34.87 & 37.89 & 37.92 & 37.78 & 37.99 & 37.98 & 37.97 & 38.06 
\\
\hline
\end{tabular*}
\end{center}

\vspace{-2mm}
\end{table}

\begin{figure}[htbp]
\scriptsize
\centering
\centerline{
\includegraphics[scale =0.56]{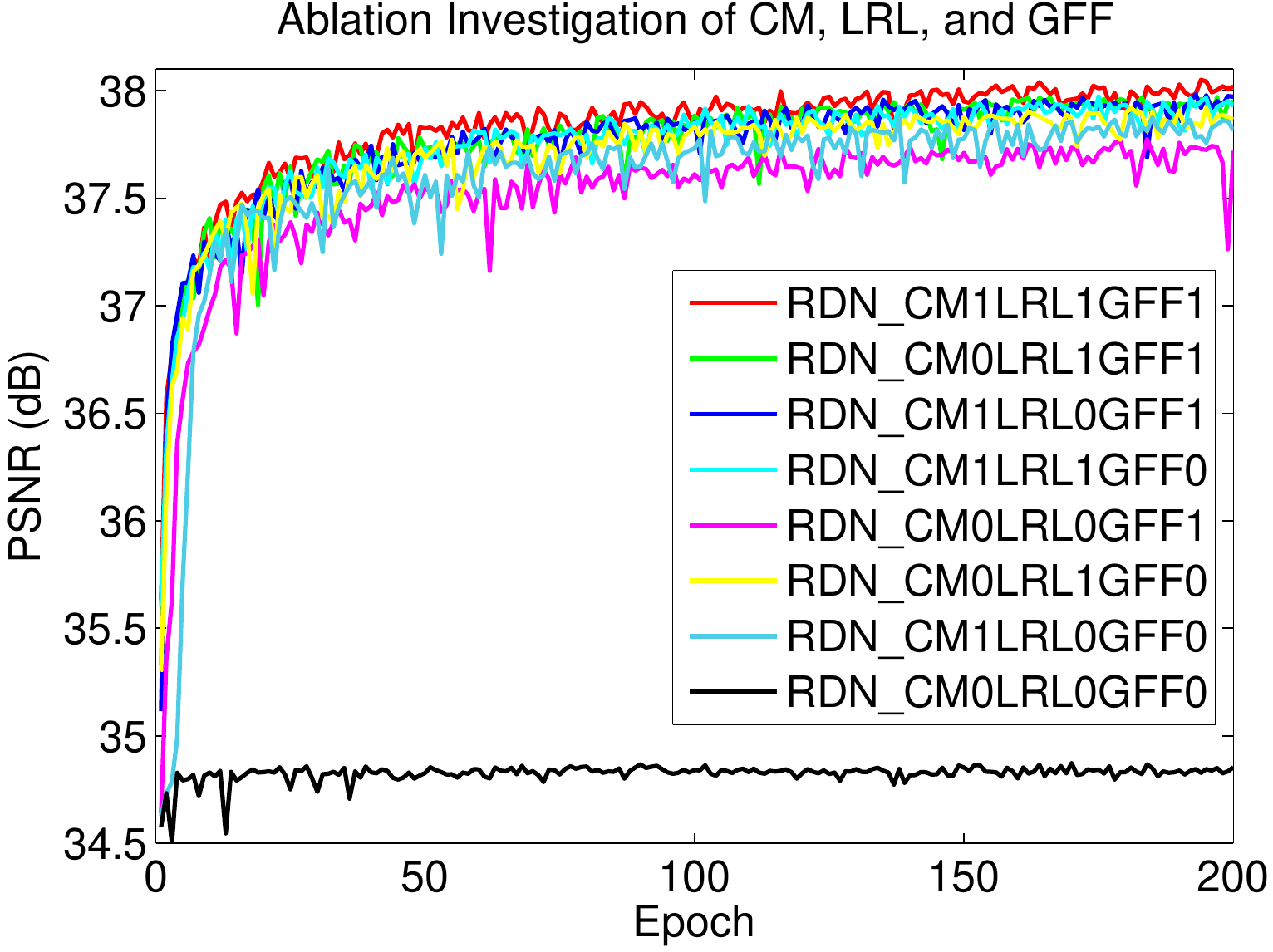}}
\vspace{-3mm}
\caption{Convergence analysis on CM, LRL, and GFF. The curves for each combination is based on the PSNR on Set5 ($\times2$) in 200 epochs.}  
\label{fig:study_CM_LRL_GFF}
\vspace{-5mm}
\end{figure}

\subsection{Ablation Investigation}
\label{subsec:ablation}
Table~\ref{tab:results_ablation} shows the ablation investigation on the effects of contiguous memory (CM), local residual learning (LRL), and global feature fusion (GFF). The eight networks have the same RDB number (D = 20), Conv number (C = 6) per RDB, and growth rate (G = 32). We find that local feature fusion (LFF) is needed to train these networks properly, so LFF isn't removed by default. The baseline (denote as RDN\_CM0LRL0GFF0) is obtained without CM, LRL, or GFF and performs very poorly (PSNR = 34.87 dB). This is caused by the difficulty of training~\cite{dong2016image} and also demonstrates that stacking many basic dense blocks~\cite{huang2017densely} in a very deep network would not result in better performance.      

We then add one of CM, LRL, or GFF to the baseline, resulting in RDN\_CM1LRL0GFF0, RDN\_CM0LRL1GFF0, and RDN\_CM0LRL0GFF1 respectively (from 2$^{nd}$ to 4$^{th}$ combination in Table~\ref{tab:results_ablation}). We can validate that each component can efficiently improve the performance of the baseline. This is mainly because each component contributes to the flow of information and gradient. 

We further add two components to the baseline, resulting in RDN\_CM1LRL1GFF0, RDN\_CM1LRL0GFF1, and RDN\_CM0LRL1GFF1 respectively (from 5$^{th}$ to 7$^{th}$ combination in Table~\ref{tab:results_ablation}). It can be seen that two components would perform better than only one component. Similar phenomenon can be seen when we use these three components simultaneously (denote as RDN\_CM1LRL1GFF1). RDN using three components performs the best.

We also visualize the convergence process of these eight combinations in Figure~\ref{fig:study_CM_LRL_GFF}. The convergence curves are consistent with the analyses above and show that CM, LRL, and GFF can further stabilize the training process without obvious performance drop. These quantitative and visual analyses demonstrate the effectiveness and benefits of our proposed CM, LRL, and GFF.


\begin{table}[t]
\scriptsize
\centering
\begin{center}
\caption{Parameter number, PSNR, and test time comparisons. The PSNR values are based on Set14 with \yulun{Bicubic} (\textbf{BI}) degradation model ($\times2$).}
\label{tab:res_modelsize_performance_time}
\vspace{-3mm}
\begin{tabular*}{82.2mm}{@{\extracolsep{-0.928mm}}|c|c|c|c|c|c|c|c|c|c|c|c|c|c|c|c|c|}
\hline
\multirow{2}{*}{Methods} &  LapSRN &  DRRN &  MemNet & MDSR & EDSR & RDN 
\\
 & \cite{lai2017deep} & \cite{tai2017image} & \cite{tai2017memnet} & \cite{lim2017enhanced} & \cite{lim2017enhanced} & (ours)
\\
\hline
\# param.
& 812K & 297K & 677K & 8M & 43M & 22M  
\\
\hline
PSNR (dB)
& 33.08 & 33.23 & 33.28 & 33.85 & 33.92 & 34.14                              
\\
\hline
Time (s)
& 0.10 & 17.81 & 13.84 & 0.53 & 1.64 & 1.56                              
\\
\hline          
\end{tabular*}
\end{center}
\vspace{-4mm}
\end{table}

\begin{figure}[t]
\scriptsize
\centering
\centerline{
\includegraphics[scale =0.55]{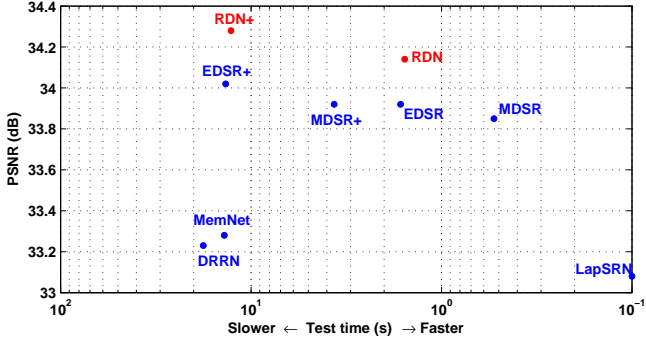}}
\vspace{-3mm}
\caption{PSNR and test time on Set14 with \textbf{BI} model ($\times2$).}  
\label{fig:psnr_time_set14x2}
\vspace{-4mm}
\end{figure}

\begin{table*}[thbp]
\scriptsize
\center
\begin{center}
\caption{Quantitative results with BI degradation model. Best and second best results are \textbf{highlighted} and \underline{underlined}.}
\label{tab:results_BI_5sets}
\vspace{-3mm}
\begin{tabular}{|l|c|c|c|c|c|c|c|c|c|c|c|}
\hline
\multirow{2}{*}{Method} & \multirow{2}{*}{Scale} &  \multicolumn{2}{c|}{Set5} &  \multicolumn{2}{c|}{Set14} &  \multicolumn{2}{c|}{B100} &  \multicolumn{2}{c|}{Urban100} &  \multicolumn{2}{c|}{Manga109}  
\\
\cline{3-12}
&  & PSNR & SSIM & PSNR & SSIM & PSNR & SSIM & PSNR & SSIM & PSNR & SSIM 
\\
\hline
\hline
Bicubic & $\times$2 
& 33.66
 & 0.9299
  & 30.24
   & 0.8688
    & 29.56
     & 0.8431
      & 26.88
       & 0.8403
        & 30.80
         & 0.9339
                  
\\
SRCNN~\cite{dong2016image} & $\times$2 
& 36.66
 & 0.9542
  & 32.45
   & 0.9067
    & 31.36
     & 0.8879
      & 29.50
       & 0.8946
        & 35.60
         & 0.9663
                   
\\
FSRCNN~\cite{dong2016accelerating} & $\times$2 
& 37.05
 & 0.9560
  & 32.66
   & 0.9090
    & 31.53
     & 0.8920
      & 29.88
       & 0.9020
        & 36.67
         & 0.9710
                   
\\
VDSR~\cite{kim2016accurate} & $\times$2 
& 37.53
 & 0.9590
  & 33.05
   & 0.9130
    & 31.90
     & 0.8960
      & 30.77
       & 0.9140
        & 37.22
         & 0.9750
                   
\\
LapSRN~\cite{lai2017deep} & $\times$2 
& 37.52
 & 0.9591
  & 33.08
   & 0.9130
    & 31.08
     & 0.8950
      & 30.41
       & 0.9101
        & 37.27
         & 0.9740
                   
\\
MemNet~\cite{tai2017memnet} & $\times$2 
& 37.78
 & 0.9597
  & 33.28
   & 0.9142
    & 32.08
     & 0.8978
      & 31.31
       & 0.9195
        & 37.72
         & 0.9740
                   
\\
\yulun{DPDNN~\cite{dong2019denoising}} & $\times$2 
& \yulun{37.75}
 & \yulun{0.9600}
  & \yulun{33.30}
   & \yulun{0.9150}
    & \yulun{32.09}
     & \yulun{0.8990}
      & \yulun{31.50}
       & \yulun{0.9220}
        & \yulun{N/A}
         & \yulun{N/A}                  
\\
EDSR~\cite{lim2017enhanced} & $\times$2 
& 38.11
 & 0.9602
  & 33.92
   & 0.9195
    & 32.32
     & 0.9013
      & 32.93
       & 0.9351
        & 39.10
         & 0.9773
                   
\\
SRMDNF~\cite{zhang2018learning} & $\times$2 
& 37.79
 & 0.9601
  & 33.32
   & 0.9159
    & 32.05
     & 0.8985
      & 31.33
       & 0.9204
        & 38.07
         & 0.9761
                   
\\
D-DBPN~\cite{haris2018deep} & $\times$2 
& 38.09
 & 0.9600
  & 33.85
   & 0.9190
    & 32.27
     & 0.9000
      & 32.55
       & 0.9324
        & 38.89
         & 0.9775        
\\
         
RDN (ours) & $\times$2 
& \underline{38.30}
 & \underline{0.9617}
  & \underline{34.14}
   & \underline{0.9235}
    & \underline{32.41}
     & \underline{0.9025}
      & \underline{33.17}
       & \underline{0.9377}
        & \underline{39.60}
         & \underline{0.9791}

\\
RDN+ (ours) & $\times$2 
& \textbf{38.34}
 & \textbf{0.9618}
  & \textbf{34.28}
   & \textbf{0.9241}
    & \textbf{32.46}
     & \textbf{0.9030}
      & \textbf{33.36}
       & \textbf{0.9388}
        & \textbf{39.74}
         & \textbf{0.9794}

\\
\hline
\hline
Bicubic & $\times$3 
& 30.39
 & 0.8682
  & 27.55
   & 0.7742
    & 27.21
     & 0.7385
      & 24.46
       & 0.7349
        & 26.95
         & 0.8556
                  
\\
SRCNN~\cite{dong2016image} & $\times$3
& 32.75
 & 0.9090
  & 29.30
   & 0.8215
    & 28.41
     & 0.7863
      & 26.24
       & 0.7989
        & 30.48
         & 0.9117
                    
\\
FSRCNN~\cite{dong2016accelerating} & $\times$3 
& 33.18
 & 0.9140
  & 29.37
   & 0.8240
    & 28.53
     & 0.7910
      & 26.43
       & 0.8080
        & 31.10
         & 0.9210
                   
\\
VDSR~\cite{kim2016accurate} & $\times$3 
& 33.67
 & 0.9210
  & 29.78
   & 0.8320
    & 28.83
     & 0.7990
      & 27.14
       & 0.8290
        & 32.01
         & 0.9340
                   
\\
LapSRN~\cite{lai2017deep} & $\times$3 
& 33.82
 & 0.9227
  & 29.87
   & 0.8320
    & 28.82
     & 0.7980
      & 27.07
       & 0.8280
        & 32.21
         & 0.9350
                   
\\
MemNet~\cite{tai2017memnet} & $\times$3 
& 34.09
 & 0.9248
  & 30.00
   & 0.8350
    & 28.96
     & 0.8001
      & 27.56
       & 0.8376
        & 32.51
         & 0.9369
                   
\\
\yulun{DPDNN~\cite{dong2019denoising}} & $\times$3 
& \yulun{33.93}
 & \yulun{0.9240}
  & \yulun{30.02}
   & \yulun{0.8360}
    & \yulun{29.00}
     & \yulun{0.8010}
      & \yulun{27.61}
       & \yulun{0.8420}
        & \yulun{N/A}
         & \yulun{N/A}                  
\\
EDSR~\cite{lim2017enhanced} & $\times$3 
& 34.65
 & 0.9280
  & 30.52
   & 0.8462
    & 29.25
     & 0.8093
      & 28.80
       & 0.8653
        & 34.17
         & 0.9476
                   
\\
SRMDNF~\cite{zhang2018learning} & $\times$3 
& 34.12
 & 0.9254
  & 30.04
   & 0.8382
    & 28.97
     & 0.8025
      & 27.57
       & 0.8398
        & 33.00
         & 0.9403
                   
\\
%
RDN (ours) & $\times$3 
& \underline{34.78}
 & \underline{0.9299}
  & \underline{30.63}
   & \underline{0.8477}
    & \underline{29.33}
     & \underline{0.8107}
      & \underline{29.02}
       & \underline{0.8695}
        & \underline{34.58}
         & \underline{0.9502}

\\
RDN+ (ours) & $\times$3 
& \textbf{34.84}
 & \textbf{0.9303}
  & \textbf{30.74}
   & \textbf{0.8495}
    & \textbf{29.38}
     & \textbf{0.8115}
      & \textbf{29.18}
       & \textbf{0.8718}
        & \textbf{34.81}
         & \textbf{0.9512}
         
\\
\hline
\hline
Bicubic & $\times$4 
& 28.42
 & 0.8104
  & 26.00
   & 0.7027
    & 25.96
     & 0.6675
      & 23.14
       & 0.6577
        & 24.89
         & 0.7866
                  
\\
SRCNN~\cite{dong2016image} & $\times$4 
& 30.48
 & 0.8628
  & 27.50
   & 0.7513
    & 26.90
     & 0.7101
      & 24.52
       & 0.7221
        & 27.58
         & 0.8555
                   
\\
FSRCNN~\cite{dong2016accelerating} & $\times$4 
& 30.72
 & 0.8660
  & 27.61
   & 0.7550
    & 26.98
     & 0.7150
      & 24.62
       & 0.7280
        & 27.90
         & 0.8610
                   
\\
VDSR~\cite{kim2016accurate} & $\times$4 
& 31.35
 & 0.8830
  & 28.02
   & 0.7680
    & 27.29
     & 0.0726
      & 25.18
       & 0.7540
        & 28.83
         & 0.8870
                   
\\
LapSRN~\cite{lai2017deep} & $\times$4 
& 31.54
 & 0.8850
  & 28.19
   & 0.7720
    & 27.32
     & 0.7270
      & 25.21
       & 0.7560
        & 29.09
         & 0.8900
                   
\\
MemNet~\cite{tai2017memnet} & $\times$4 
& 31.74
 & 0.8893
  & 28.26
   & 0.7723
    & 27.40
     & 0.7281
      & 25.50
       & 0.7630
        & 29.42
         & 0.8942
                   
\\
\yulun{DPDNN~\cite{dong2019denoising}} & $\times$4 
& \yulun{31.72}
 & \yulun{0.8890}
  & \yulun{28.28}
   & \yulun{0.7730}
    & \yulun{27.44}
     & \yulun{0.7290}
      & \yulun{25.53}
       & \yulun{0.7680}
        & \yulun{N/A}
         & \yulun{N/A}                  
\\
SRDenseNet~\cite{tong2017image} & $\times$4 
& 32.02
 & 0.8930
  & 28.50
   & 0.7780
    & 27.53
     & 0.7337
      & 26.05
       & 0.7819
        & N/A
         & N/A
                   
\\
EDSR~\cite{lim2017enhanced} & $\times$4 
& 32.46
 & 0.8968
  & 28.80
   & 0.7876
    & 27.71
     & 0.7420
      & 26.64
       & 0.8033
        & 31.02
         & 0.9148
                   
\\
SRMDNF~\cite{zhang2018learning} & $\times$4 
& 31.96
 & 0.8925
  & 28.35
   & 0.7787
    & 27.49
     & 0.7337
      & 25.68
       & 0.7731
        & 30.09
         & 0.9024
                   
\\
D-DBPN~\cite{haris2018deep} & $\times$4 
& 32.47
 & 0.8980
  & 28.82
   & 0.7860
    & 27.72
     & 0.7400
      & 26.38
       & 0.7946
        & 30.91
         & 0.9137
         
\\
         
RDN (ours) & $\times$4 
& \underline{32.61}
 & \underline{0.8999}
  & \underline{28.93}
   & \underline{0.7894}
    & \underline{27.80}
     & \underline{0.7436}
      & \underline{26.85}
       & \underline{0.8089}
        & \underline{31.45}
         & \underline{0.9187}

\\
RDN+ (ours) & $\times$4 
& \textbf{32.69}
 & \textbf{0.9007}
  & \textbf{29.01}
   & \textbf{0.7909}
    & \textbf{27.85}
     & \textbf{0.7447}
      & \textbf{27.01}
       & \textbf{0.8120}
        & \textbf{31.74}
         & \textbf{0.9208}
              
\\
\hline
\hline
Bicubic & $\times$8 
& 24.40
 & 0.6580
  & 23.10
   & 0.5660
    & 23.67
     & 0.5480
      & 20.74
       & 0.5160
        & 21.47
         & 0.6500
                 
\\
SRCNN~\cite{dong2016image} & $\times$8 
& 25.33
 & 0.6900
  & 23.76
   & 0.5910
    & 24.13
     & 0.5660
      & 21.29
       & 0.5440
        & 22.46
         & 0.6950
                   
\\
FSRCNN~\cite{dong2016accelerating} & $\times$8 
& 20.13
 & 0.5520
  & 19.75
   & 0.4820
    & 24.21
     & 0.5680
      & 21.32
       & 0.5380
        & 22.39
         & 0.6730
                   
\\
SCN~\cite{wang2015deep} & $\times$8 
& 25.59
 & 0.7071
  & 24.02
   & 0.6028
    & 24.30
     & 0.5698
      & 21.52
       & 0.5571
        & 22.68
         & 0.6963

\\
VDSR~\cite{kim2016accurate} & $\times$8 
& 25.93
 & 0.7240
  & 24.26
   & 0.6140
    & 24.49
     & 0.5830
      & 21.70
       & 0.5710
        & 23.16
         & 0.7250
                   
\\   
LapSRN~\cite{lai2017deep} & $\times$8 
& 26.15
 & 0.7380
  & 24.35
   & 0.6200
    & 24.54
     & 0.5860
      & 21.81
       & 0.5810
        & 23.39
         & 0.7350
                   
\\
MemNet~\cite{tai2017memnet} & $\times$8 
& 26.16
 & 0.7414
  & 24.38
   & 0.6199
    & 24.58
     & 0.5842
      & 21.89
       & 0.5825
        & 23.56
         & 0.7387

\\
MSLapSRN~\cite{lai2018MSLapSRN} & $\times$8 
& 26.34
 & 0.7558
  & 24.57
   & 0.6273
    & 24.65
     & 0.5895
      & 22.06
       & 0.5963
        & 23.90
         & 0.7564
                   
\\
EDSR~\cite{lim2017enhanced} & $\times$8 
& 26.96
 & 0.7762
  & 24.91
   & 0.6420
    & 24.81
     & 0.5985
      & 22.51
       & 0.6221
        & 24.69
         & 0.7841
                   
\\
D-DBPN~\cite{haris2018deep} & $\times$8 
& 27.21
 & 0.7840
  & 25.13
   & 0.6480
    & 24.88
     & 0.6010
      & 22.73
       & 0.6312
        & 25.14
         & 0.7987
         
\\
RDN (ours) & $\times$8 
& \underline{27.23}
 & \underline{0.7854}
  & \underline{25.25}
   & \underline{0.6505}
    & \underline{24.91}
     & \underline{0.6032}
      & \underline{22.83}
       & \underline{0.6374}
        & \underline{25.14}
         & \underline{0.7994}

\\
RDN+ (ours) & $\times$8 
& \textbf{27.40}
 & \textbf{0.7900}
  & \textbf{25.38}
   & \textbf{0.6541}
    & \textbf{25.01}
     & \textbf{0.6057}
      & \textbf{23.04}
       & \textbf{0.6439}
        & \textbf{25.48}
         & \textbf{0.8058}
           
\\
\hline             
\end{tabular}
\end{center}
\vspace{-6mm}
\end{table*}

\subsection{Model Size, Performance, and Test Time}
We also compare the model size, performance, and test time with other methods on Set14 ($\times2$) in Table~\ref{tab:res_modelsize_performance_time}. Compared with EDSR, our RDN has half less amount of parameter and obtains better results. Although our RDN has more parameters than that of other methods, RDN achieves comparable (\eg, MDSR) or even less test time (\eg, MemNet). We further visualize the performance and test time comparison in Figure~\ref{fig:psnr_time_set14x2}. We can see that our RDN achieves good trade-off between the performance and running time. 
\vspace{-3mm}

\begin{figure*}[htpb]
\scriptsize
\centering
\begin{tabular}{cc}
\hspace{-0.4cm}
\begin{adjustbox}{valign=t}
\begin{tabular}{c}
\includegraphics[width=0.1955\textwidth]{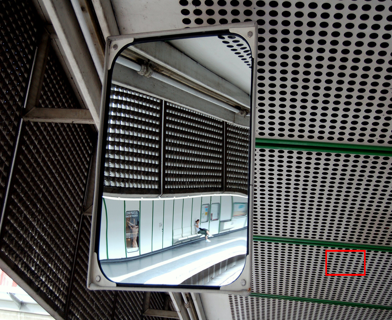}
\\
Urban100: img\_004 ($\times$4)
\end{tabular}
\end{adjustbox}
\hspace{-0.46cm}
\begin{adjustbox}{valign=t}
\begin{tabular}{cccccc}
\includegraphics[width=0.138\textwidth]{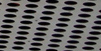} \hspace{-4mm} &
\includegraphics[width=0.138\textwidth]{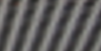} \hspace{-4mm} &
\includegraphics[width=0.138\textwidth]{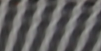} \hspace{-4mm} &
\includegraphics[width=0.138\textwidth]{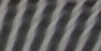} \hspace{-4mm} &
\includegraphics[width=0.138\textwidth]{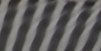} \hspace{-4mm} 
\\
GT \hspace{-4mm} &
Bicubic \hspace{-4mm} &
SRCNN~\cite{dong2016image} \hspace{-4mm} &
FSRCNN~\cite{dong2016accelerating} \hspace{-4mm} &
VDSR~\cite{kim2016accurate}
\\
\includegraphics[width=0.138\textwidth]{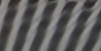} \hspace{-4mm} &
\includegraphics[width=0.138\textwidth]{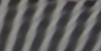} \hspace{-4mm} &
\includegraphics[width=0.138\textwidth]{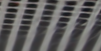} \hspace{-4mm} &
\includegraphics[width=0.138\textwidth]{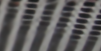} \hspace{-4mm} &
\includegraphics[width=0.138\textwidth]{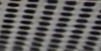} \hspace{-4mm}  
\\ 
MemNet~\cite{tai2017memnet} \hspace{-4mm} &
LapSRN~\cite{lai2017deep} \hspace{-4mm} &
EDSR~\cite{lim2017enhanced} \hspace{-4mm} &
D-DBPN~\cite{haris2018deep}  \hspace{-4mm} &
RDN (ours)  \hspace{-4.2mm} 
\\
\end{tabular}
\end{adjustbox}
\vspace{2mm}
\\
\hspace{-0.4cm}
\begin{adjustbox}{valign=t}
\begin{tabular}{c}
\includegraphics[width=0.1955\textwidth]{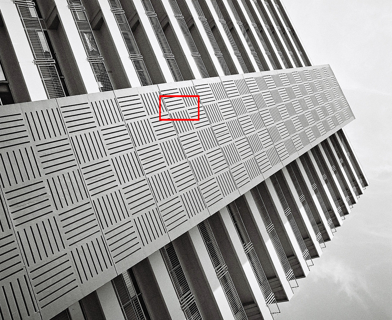}
\\
Urban100: img\_092 ($\times$4)
\end{tabular}
\end{adjustbox}
\hspace{-0.46cm}
\begin{adjustbox}{valign=t}
\begin{tabular}{cccccc}
\includegraphics[width=0.138\textwidth]{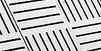} \hspace{-4mm} &
\includegraphics[width=0.138\textwidth]{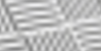} \hspace{-4mm} &
\includegraphics[width=0.138\textwidth]{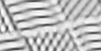} \hspace{-4mm} &
\includegraphics[width=0.138\textwidth]{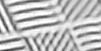} \hspace{-4mm} &
\includegraphics[width=0.138\textwidth]{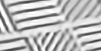} \hspace{-4mm} 
\\
GT \hspace{-4mm} &
Bicubic \hspace{-4mm} &
SRCNN~\cite{dong2016image} \hspace{-4mm} &
FSRCNN~\cite{dong2016accelerating} \hspace{-4mm} &
VDSR~\cite{kim2016accurate}
\\
\includegraphics[width=0.138\textwidth]{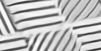} \hspace{-4mm} &
\includegraphics[width=0.138\textwidth]{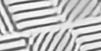} \hspace{-4mm} &
\includegraphics[width=0.138\textwidth]{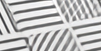} \hspace{-4mm} &
\includegraphics[width=0.138\textwidth]{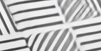} \hspace{-4mm} &
\includegraphics[width=0.138\textwidth]{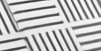} \hspace{-4mm}  
\\ 
MemNet~\cite{tai2017memnet} \hspace{-4mm} &
LapSRN~\cite{lai2017deep} \hspace{-4mm} &
EDSR~\cite{lim2017enhanced} \hspace{-4mm} &
D-DBPN~\cite{haris2018deep}  \hspace{-4mm} &
RDN (ours)  \hspace{-4.2mm} 
\\
\end{tabular}
\end{adjustbox}
\vspace{2mm}
\\
\hspace{-0.4cm}
\begin{adjustbox}{valign=t}
\begin{tabular}{c}
\includegraphics[width=0.1955\textwidth]{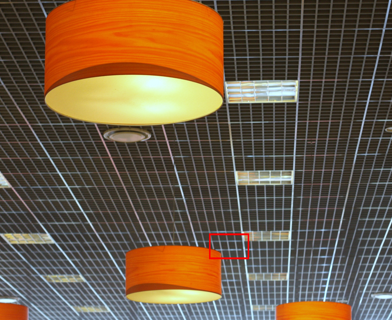}
\\
Urban100: img\_044 ($\times$8)
\end{tabular}
\end{adjustbox}
\hspace{-0.46cm}
\begin{adjustbox}{valign=t}
\begin{tabular}{cccccc}
\includegraphics[width=0.138\textwidth]{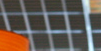} \hspace{-4mm} &
\includegraphics[width=0.138\textwidth]{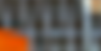} \hspace{-4mm} &
\includegraphics[width=0.138\textwidth]{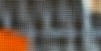} \hspace{-4mm} &
\includegraphics[width=0.138\textwidth]{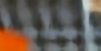} \hspace{-4mm} &
\includegraphics[width=0.138\textwidth]{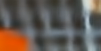} \hspace{-4mm} 
\\
GT \hspace{-4mm} &
Bicubic \hspace{-4mm} &
SRCNN~\cite{dong2016image} \hspace{-4mm} &
SCN~\cite{wang2015deep} \hspace{-4mm} &
VDSR~\cite{kim2016accurate}
\\
\includegraphics[width=0.138\textwidth]{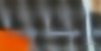} \hspace{-4mm} &
\includegraphics[width=0.138\textwidth]{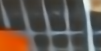} \hspace{-4mm} &
\includegraphics[width=0.138\textwidth]{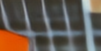} \hspace{-4mm} &
\includegraphics[width=0.138\textwidth]{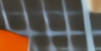} \hspace{-4mm} &
\includegraphics[width=0.138\textwidth]{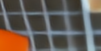} \hspace{-4mm}  
\\ 
MemNet~\cite{tai2017memnet} \hspace{-4mm} &
MSLapSRN~\cite{lai2018MSLapSRN} \hspace{-4mm} &
EDSR~\cite{lim2017enhanced} \hspace{-4mm} &
D-DBPN~\cite{haris2018deep}  \hspace{-4mm} &
RDN (ours) \hspace{-4.2mm}
\\
\end{tabular}
\end{adjustbox}
\vspace{2mm}
\\
\hspace{-0.4cm}
\begin{adjustbox}{valign=t}
\begin{tabular}{c}
\includegraphics[width=0.1955\textwidth]{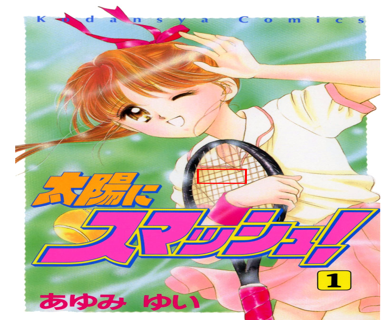}
\\
Manga109: TaiyouNiSmash ($\times$8)
\end{tabular}
\end{adjustbox}
\hspace{-0.46cm}
\begin{adjustbox}{valign=t}
\begin{tabular}{cccccc}
\includegraphics[width=0.138\textwidth]{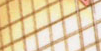} \hspace{-4mm} &
\includegraphics[width=0.138\textwidth]{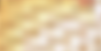} \hspace{-4mm} &
\includegraphics[width=0.138\textwidth]{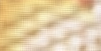} \hspace{-4mm} &
\includegraphics[width=0.138\textwidth]{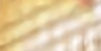} \hspace{-4mm} &
\includegraphics[width=0.138\textwidth]{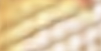} \hspace{-4mm} 
\\
GT \hspace{-4mm} &
Bicubic \hspace{-4mm} &
SRCNN~\cite{dong2016image} \hspace{-4mm} &
SCN~\cite{wang2015deep} \hspace{-4mm} &
VDSR~\cite{kim2016accurate}
\\
\includegraphics[width=0.138\textwidth]{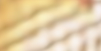} \hspace{-4mm} &
\includegraphics[width=0.138\textwidth]{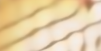} \hspace{-4mm} &
\includegraphics[width=0.138\textwidth]{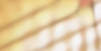} \hspace{-4mm} &
\includegraphics[width=0.138\textwidth]{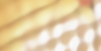} \hspace{-4mm} &
\includegraphics[width=0.138\textwidth]{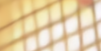} \hspace{-4mm}  
\\ 
MemNet~\cite{tai2017memnet} \hspace{-4mm} &
MSLapSRN~\cite{lai2018MSLapSRN} \hspace{-4mm} &
EDSR~\cite{lim2017enhanced} \hspace{-4mm} &
D-DBPN~\cite{haris2018deep}  \hspace{-4mm} &
RDN (ours)  \hspace{-4.2mm} 
\\
\end{tabular}
\end{adjustbox}
\end{tabular}
\vspace{-2mm}
\caption{Image super-resolution results (\textbf{BI} degradation model) with scaling factors $s$ = 4 (first two rows) and $s$ = 8 (last two rows).}
\label{fig:result_SR_RGB_X4X8}
\vspace{-3mm}
\end{figure*}

\section{Experimental Results}
\label{sec:results}
The source code and models of the proposed method can be downloaded at \href{https://github.com/yulunzhang/RDN}{https://github.com/yulunzhang/RDN}.

\subsection{Settings}
\label{subsec:settings}
\yulun{Here we provide details of experimental settings, such as training/testing data for each tasks.}

\yulun{\textbf{Training Data}. Recently, Timofte et al. have released a high-quality (2K resolution) dataset DIV2K for image restoration applications~\cite{timofte2017ntire}. DIV2K consists of 800 training images, 100 validation images, and 100 test images. We use DIV2K as training data for image SR (except for Bicubic degradation model), color and gray image denoising, CAR, and deblurring. For image SR with Bicubic degradation (\textbf{BI}), some compared methods (\eg, D-DBPN~\cite{haris2018deep}) further use Flickr2K~\cite{lim2017enhanced} as well as DIV2K for training. We also train RDN by using larger training data to investigate whether RDN can further improve performance.}

\yulun{\textbf{Testing Data and Metrics.} For testing, we use five standard benchmark datasets: Set5~\cite{bevilacqua2012low}, Set14~\cite{zeyde2012single}, B100~\cite{martin2001database}, Urban100~\cite{huang2015single}, and Manga109~\cite{matsui2017sketch} for image SR. We use Kodak24~(\href{http://r0k.us/graphics/kodak/}{http://r0k.us/graphics/kodak/}), BSD68~\cite{martin2001database}, and Urban100~\cite{huang2015single} for color and gray image DN. LIVE1~\cite{sheikh2005live} and Classic5~\cite{foi2007pointwise} are used for image CAR. We use McMaster18~\cite{zhang2017learning}, Kodak24, and Urban100 as testing data for image deblurring. The quantitative results are evaluated with PSNR and SSIM~\cite{wang2004image} on Y channel ({i.e.}, luminance) of transformed YCbCr space.}

\textbf{Degradation Models.} \yulun{For image SR,} in order to fully demonstrate the effectiveness of our proposed RDN, we use three degradation models to simulate LR images for image SR. The first one is bicubic downsampling by adopting the Matlab function \textit{imresize} with the option \textit{bicubic} (denote as \textbf{BI} for short). We use \textbf{BI} model to simulate LR images with scaling factor $\times2$, $\times3$, $\times4$, and $\times8$. Similar to~\cite{zhang2017learning}, the second one is to blur HR image by Gaussian kernel of size $7\times7$ with standard deviation 1.6. The blurred image is then downsampled with scaling factor $\times3$ (denote as \textbf{BD} for short). We further produce LR image in a more challenging way. We first bicubic downsample HR image with scaling factor $\times3$ and then add Gaussian noise with noise level 30 (denote as \textbf{DN} for short). \yulun{For color and gray image denoising, we add Gaussian noise with noise level $\sigma$ to the ground truth to obtain the noisy inputs. For image CAR, we use Matlab JPEG encoder~\cite{jancsary2012loss} to generate compressed inputs. For image deburring, the commonly-used 25$\times$25 Gaussian blur kernel of standard deviation 1.6 is used to blur images first. Then, additive Gaussian noise ($\sigma=2$) is added to the blurry images to obtain the final inputs.}

\textbf{Training Setting.}
Following settings of~\cite{lim2017enhanced}, in each training batch, we randomly extract 16 LQ RGB patches with the size of $48\times48$ as inputs \yulun{for image SR, DN, CAR, and deblurring}. We randomly augment the patches by flipping horizontally or vertically and rotating 90$^{\circ}$. 1,000 iterations of back-propagation constitute an epoch. We implement our RDN with the Torch7 framework and update it with Adam optimizer~\cite{kingma2014adam}. The learning rate is initialized to 10$^{-4}$ for all layers and decreases half for every 200 epochs. Training a RDN roughly takes 1 day with a Titan Xp GPU for 200 epochs.

\begin{table*}[htbp]
\scriptsize
\centering
\begin{center}
\caption{Benchmark results with \textbf{BD} and \textbf{DN} degradation models. Average PSNR/SSIM values for scaling factor $\times3$.} 
\label{tab:results_BD_DN_5sets}
\vspace{-3mm}
\begin{tabular*}{163mm}{@{\extracolsep{-0.928mm}}|c|c|c|c|c|c|c|c|c|c|c|c|c|c|c|c|c|}
\hline
\multirow{2}{*}{Dataset} & \multirow{2}{*}{Model} &  \multirow{2}{*}{Bicubic} & SRCNN &  FSRCNN &  VDSR &  IRCNN\_G &  IRCNN\_C & RDN & RDN+ 
\\
&  &  & \cite{dong2016image} & \cite{dong2016accelerating} & \cite{kim2016accurate} & \cite{zhang2017learning} & \cite{zhang2017learning} & (ours) & (ours)    
\\
\hline
\hline
\multirow{2}{*}{Set5}
& \textbf{BD} 
& 28.78/0.8308
  & 32.05/0.8944
   & 26.23/0.8124
    & 33.25/0.9150
     & 33.38/0.9182
      & 33.17/0.9157
       & 34.58/0.9280
        & \textbf{34.70}/\textbf{0.9289}
                            
\\
& \textbf{DN} 
& 24.01/0.5369
  & 25.01/0.6950
   & 24.18/0.6932
    & 25.20/0.7183
     & 25.70/0.7379
      & 27.48/0.7925
       & 28.47/0.8151
        & \textbf{28.55}/\textbf{0.8173}
                              
\\
\hline 
\hline
\multirow{2}{*}{Set14}
& \textbf{BD} 
& 26.38/0.7271
  & 28.80/0.8074
   & 24.44/0.7106
    & 29.46/0.8244
     & 29.63/0.8281
      & 29.55/0.8271
       & 30.53/0.8447
        & \textbf{30.64}/\textbf{0.8463}
                            
\\
& \textbf{DN} 
& 22.87/0.4724
  & 23.78/0.5898
   & 23.02/0.5856
    & 24.00/0.6112
     & 24.45/0.6305
      & 25.92/0.6932
       & 26.60/0.7101
        & \textbf{26.67}/\textbf{0.7117}

\\
\hline
\hline
\multirow{2}{*}{B100}
& \textbf{BD} 
& 26.33/0.6918
  & 28.13/0.7736
   & 24.86/0.6832
    & 28.57/0.7893
     & 28.65/0.7922
      & 28.49/0.7886
       & 29.23/0.8079
        & \textbf{29.30}/\textbf{0.8093}
                            
\\
& \textbf{DN}
& 22.92/0.4449
  & 23.76/0.5538
   & 23.41/0.5556
    & 24.00/0.5749
     & 24.28/0.5900
      & 25.55/0.6481
       & 25.93/0.6573
        & \textbf{25.97}/\textbf{0.6587}

\\
\hline
\hline
\multirow{2}{*}{Urban100}
& \textbf{BD} 
& 23.52/0.6862
  & 25.70/0.7770
   & 22.04/0.6745
    & 26.61/0.8136
     & 26.77/0.8154
      & 26.47/0.8081
       & 28.46/0.8582
        & \textbf{28.67}/\textbf{0.8612}
                            
\\
& \textbf{DN} 
& 21.63/0.4687
  & 21.90/0.5737
   & 21.15/0.5682
    & 22.22/0.6096
     & 22.90/0.6429
      & 23.93/0.6950
       & 24.92/0.7364
        & \textbf{25.05}/\textbf{0.7399}
                              
\\
\hline
\hline
\multirow{2}{*}{Manga109}
& \textbf{BD} 
& 25.46/0.8149
  & 29.47/0.8924
   & 23.04/0.7927
    & 31.06/0.9234
     & 31.15/0.9245
      & 31.13/0.9236
       & 33.97/0.9465
        & \textbf{34.34}/\textbf{0.9483}
                            
\\
& \textbf{DN} 
& 23.01/0.5381
  & 23.75/0.7148
   & 22.39/0.7111
    & 24.20/0.7525
     & 24.88/0.7765
      & 26.07/0.8253
       & 28.00/0.8591
        & \textbf{28.18}/\textbf{0.8621}
                              
\\
\hline      
    
\end{tabular*}
\end{center}
\vspace{-3mm}
\end{table*}

\begin{figure*}[t]
\scriptsize
\centering
\begin{tabular}{cc}
\hspace{-0.4cm}
\begin{adjustbox}{valign=t}
\begin{tabular}{ccccccc}
\includegraphics[width=0.136\textwidth]{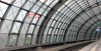} \hspace{-4.2mm} &
\includegraphics[width=0.136\textwidth]{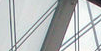} \hspace{-4.2mm} &
\includegraphics[width=0.136\textwidth]{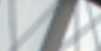} \hspace{-4.2mm} &
\includegraphics[width=0.136\textwidth]{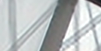} \hspace{-4.2mm} &
\includegraphics[width=0.136\textwidth]{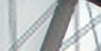} \hspace{-4.2mm} &
\includegraphics[width=0.136\textwidth]{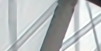} \hspace{-4.2mm} &
\includegraphics[width=0.136\textwidth]{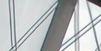} \hspace{-4.2mm} 
\\

\includegraphics[width=0.136\textwidth]{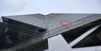} \hspace{-4.2mm} &
\includegraphics[width=0.136\textwidth]{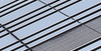} \hspace{-4.2mm} &
\includegraphics[width=0.136\textwidth]{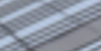} \hspace{-4.2mm} &
\includegraphics[width=0.136\textwidth]{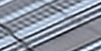} \hspace{-4.2mm} &
\includegraphics[width=0.136\textwidth]{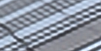} \hspace{-4.2mm} &
\includegraphics[width=0.136\textwidth]{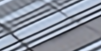} \hspace{-4.2mm} &
\includegraphics[width=0.136\textwidth]{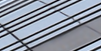} \hspace{-4.2mm} 
\\

\includegraphics[width=0.136\textwidth]{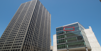} \hspace{-4.2mm} &
\includegraphics[width=0.136\textwidth]{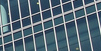} \hspace{-4.2mm} &
\includegraphics[width=0.136\textwidth]{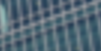} \hspace{-4.2mm} &
\includegraphics[width=0.136\textwidth]{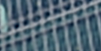} \hspace{-4.2mm} &
\includegraphics[width=0.136\textwidth]{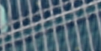} \hspace{-4.2mm} &
\includegraphics[width=0.136\textwidth]{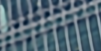} \hspace{-4.2mm} &
\includegraphics[width=0.136\textwidth]{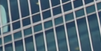} \hspace{-4.2mm} 
\\

\includegraphics[width=0.136\textwidth]{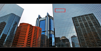} \hspace{-4.2mm} &
\includegraphics[width=0.136\textwidth]{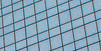} \hspace{-4.2mm} &
\includegraphics[width=0.136\textwidth]{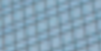} \hspace{-4.2mm} &
\includegraphics[width=0.136\textwidth]{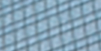} \hspace{-4.2mm} &
\includegraphics[width=0.136\textwidth]{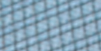} \hspace{-4.2mm} &
\includegraphics[width=0.136\textwidth]{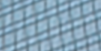} \hspace{-4.2mm} &
\includegraphics[width=0.136\textwidth]{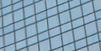} \hspace{-4.2mm} 
\\
 \hspace{-4.2mm} &
GT \hspace{-4.2mm} &
Bicubic \hspace{-4.2mm} &
SPMSR~\cite{peleg2014statistical} \hspace{-4.2mm} &
SRCNN~\cite{dong2016image} \hspace{-4.2mm} &
IRCNN~\cite{zhang2017learning} \hspace{-4.2mm} &
RDN (ours)  \hspace{-4.2mm}
\\
\end{tabular}
\end{adjustbox}

\end{tabular}
\vspace{-3mm}
\caption{Visual results using \textbf{BD} degradation model with scaling factor $\times3$.}
\label{fig:result_SR_RGB_BDX3}
\vspace{-2mm}
\end{figure*}

\begin{figure*}[htpb]
\scriptsize
\centering
\begin{tabular}{cc}
\hspace{-0.4cm}
\begin{adjustbox}{valign=t}
\begin{tabular}{ccccccc}
\includegraphics[width=0.136\textwidth]{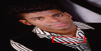} \hspace{-4.2mm} &
\includegraphics[width=0.136\textwidth]{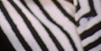} \hspace{-4.2mm} &
\includegraphics[width=0.136\textwidth]{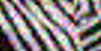} \hspace{-4.2mm} &
\includegraphics[width=0.136\textwidth]{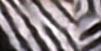} \hspace{-4.2mm} &
\includegraphics[width=0.136\textwidth]{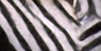} \hspace{-4.2mm} &
\includegraphics[width=0.136\textwidth]{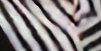} \hspace{-4.2mm} &
\includegraphics[width=0.136\textwidth]{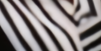} \hspace{-4.2mm} 
\\

\includegraphics[width=0.136\textwidth]{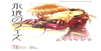} \hspace{-4.2mm} &
\includegraphics[width=0.136\textwidth]{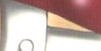} \hspace{-4.2mm} &
\includegraphics[width=0.136\textwidth]{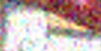} \hspace{-4.2mm} &
\includegraphics[width=0.136\textwidth]{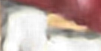} \hspace{-4.2mm} &
\includegraphics[width=0.136\textwidth]{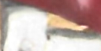} \hspace{-4.2mm} &
\includegraphics[width=0.136\textwidth]{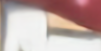} \hspace{-4.2mm} &
\includegraphics[width=0.136\textwidth]{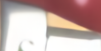} \hspace{-4.2mm} 
\\

\includegraphics[width=0.136\textwidth]{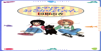} \hspace{-4.2mm} &
\includegraphics[width=0.136\textwidth]{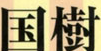} \hspace{-4.2mm} &
\includegraphics[width=0.136\textwidth]{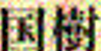} \hspace{-4.2mm} &
\includegraphics[width=0.136\textwidth]{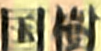} \hspace{-4.2mm} &
\includegraphics[width=0.136\textwidth]{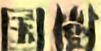} \hspace{-4.2mm} &
\includegraphics[width=0.136\textwidth]{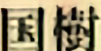} \hspace{-4.2mm} &
\includegraphics[width=0.136\textwidth]{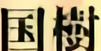} \hspace{-4.2mm} 
\\

\includegraphics[width=0.136\textwidth]{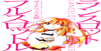} \hspace{-4.2mm} &
\includegraphics[width=0.136\textwidth]{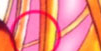} \hspace{-4.2mm} &
\includegraphics[width=0.136\textwidth]{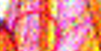} \hspace{-4.2mm} &
\includegraphics[width=0.136\textwidth]{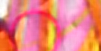} \hspace{-4.2mm} &
\includegraphics[width=0.136\textwidth]{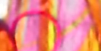} \hspace{-4.2mm} &
\includegraphics[width=0.136\textwidth]{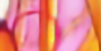} \hspace{-4.2mm} &
\includegraphics[width=0.136\textwidth]{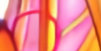} \hspace{-4.2mm} 
\\
 \hspace{-4.2mm} &
GT \hspace{-4.2mm} &
Bicubic \hspace{-4.2mm} &
SRCNN~\cite{dong2016image} \hspace{-4.2mm} &
VDSR~\cite{kim2016accurate} \hspace{-4.2mm} &
IRCNN~\cite{zhang2017learning} \hspace{-4.2mm} &
RDN (ours) \hspace{-4.2mm}
\\
\end{tabular}
\end{adjustbox}

\end{tabular}
\vspace{-3mm}
\caption{Visual results using \textbf{DN} degradation model with scaling factor $\times3$.}
\label{fig:result_SR_RGB_DNX3}
\vspace{-4mm}
\end{figure*}

\subsection{Image Super-Resolution}
\subsubsection{Results with BI Degradation Model}
\label{subsec:BI-degradation}
Simulating LR image with BI degradation model is widely used in image SR settings. For BI degradation model, we compare our RDN with state-of-the-art image SR methods: SRCNN~\cite{dong2016image}, FSRCNN~\cite{dong2016accelerating}, SCN~\cite{wang2015deep}, VDSR~\cite{kim2016accurate}, LapSRN~\cite{lai2017deep}, MemNet~\cite{tai2017memnet}, SRDenseNet~\cite{tong2017image}, MSLapSRN~\cite{lai2018MSLapSRN}, EDSR~\cite{lim2017enhanced}, SRMDNF~\cite{zhang2018learning}, D-DBPN~\cite{haris2018deep}, and \yulun{DPDNN~\cite{dong2019denoising}}. Similar to~\cite{timofte2016seven,lim2017enhanced}, we also adopt self-ensemble strategy~\cite{lim2017enhanced} to further improve our RDN and denote the self-ensembled RDN as RDN+. Here, we also additionally use Flickr2K~\cite{timofte2017ntire} as training data, which is also used in SRMDNF~\cite{zhang2018learning}, and D-DBPN~\cite{haris2018deep}. As analyzed above, a deeper and wider RDN would lead to a better performance. On the other hand, as most methods for comparison only use about 64 filters per Conv layer, we report results of RDN by using D = 16, C = 8, and G = 64 for a fair comparison. 

Table~\ref{tab:results_BI_5sets} shows quantitative comparisons for $\times2$, $\times3$, and $\times4$ SR. Results of SRDenseNet~\cite{tong2017image} are cited from their paper. When compared with persistent CNN models ( SRDenseNet~\cite{tong2017image} and MemNet~\cite{tai2017memnet}), our RDN performs the best on all datasets with all scaling factors. This indicates the better effectiveness of our residual dense block (RDB) over dense block in SRDensenet~\cite{tong2017image} and the memory block in MemNet~\cite{tai2017memnet}. When compared with the remaining models, our RDN also achieves the best average results on most datasets. Specifically, for the scaling factor $\times2$, our RDN performs the best on all datasets. EDSR~\cite{lim2017enhanced} uses far more filters (i.e., 256) per Conv layer, leading to a very wide network with a large number of parameters (i.e., 43 M). Our RDN has about half less network parameter number and achieves better performance.


In Figure~\ref{fig:result_SR_RGB_X4X8}, we show visual comparisons on scales $\times4$ and $\times8$. We observe that most of compared methods cannot recover the lost details in the LR image (\eg, ``img\_004''), even though EDSR and D-DBPN can reconstruct partial details. In contrast, our RDN can recover sharper and clearer edges, more faithful to the ground truth. In image ``img\_092'', some unwanted artifacts are generated in the degradation process. All the compared methods would fail to handle such a case, but enlarge the mistake. However, our RDN can alleviate the degradation artifacts and recover correct structures. When scaling factor goes larger (\eg, $\times8$), more structural and textural details are lost. Even we human beings can hardly distinguish the semantic content in the LR images. Most compared methods cannot recover the lost details either. However, with the usage of hierarchical features through dense feature fusion, our RDN reconstruct better visual results with clearer structures.


\vspace{-2mm}
\subsubsection{Results with BD and DN Degradation Models}
\label{subsec:BD-DN-degradation}
Following~\cite{zhang2017learning}, we also show the SR results with BD degradation model and further introduce DN degradation model. Our RDN is compared with SPMSR~\cite{peleg2014statistical}, SRCNN~\cite{dong2016image}, FSRCNN~\cite{dong2016accelerating}, VDSR~\cite{kim2016accurate}, IRCNN\_G~\cite{zhang2017learning}, and IRCNN\_C~\cite{zhang2017learning}. We re-train SRCNN, FSRCNN, and VDSR for each degradation model. Table~\ref{tab:results_BD_DN_5sets} shows the average PSNR and SSIM results on Set5, Set14, B100, Urban100, and Manga109 with scaling factor $\times3$. Our RDN and RDN+ perform the best on all the datasets with BD and DN degradation models. The performance gains over other state-of-the-art methods are consistent with the visual results in Figures~\ref{fig:result_SR_RGB_BDX3} and~\ref{fig:result_SR_RGB_DNX3}. 

For \textbf{BD} degradation model (Figure~\ref{fig:result_SR_RGB_BDX3}), the methods using interpolated LR image as input would produce noticeable artifacts and be unable to remove the blurring artifacts. In contrast, our RDN suppresses the blurring artifacts and recovers sharper edges. This comparison indicates that extracting hierarchical features from the original LR image would alleviate the blurring artifacts. It also demonstrates the strong ability of RDN for \textbf{BD} degradation model.   

For \textbf{DN} degradation model (Figure~\ref{fig:result_SR_RGB_DNX3}), where the LR image is corrupted by noise and loses some details. We observe that the noised details are hard to recovered by other methods~\cite{dong2016image,kim2016accurate,zhang2017learning}. However, our RDN can not only handle the noise efficiently, but also recover more details. This comparison indicates that RDN is applicable for jointly image denoising and SR. These results with \textbf{BD} and \textbf{DN} degradation models demonstrate the effectiveness and robustness of our RDN model. 

\begin{figure}[t]
\scriptsize
\centering
\begin{tabular}{cc}
\hspace{-0.4cm}
\begin{adjustbox}{valign=t}
\begin{tabular}{c}
\includegraphics[width=0.1420\textwidth,height=0.1402\textwidth]{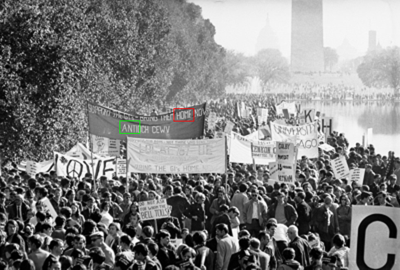}
\\
\yulun{Historical: img004}
\end{tabular}
\end{adjustbox}
\hspace{-0.46cm}
\begin{adjustbox}{valign=t}
\begin{tabular}{cccccc}
\includegraphics[width=0.108\textwidth]{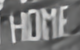} \hspace{-4mm} &
\includegraphics[width=0.108\textwidth]{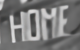} \hspace{-4mm} &
\includegraphics[width=0.108\textwidth]{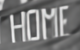} \hspace{-4mm} 
\\
\includegraphics[width=0.108\textwidth]{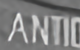} \hspace{-4mm} &
\includegraphics[width=0.108\textwidth]{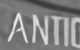} \hspace{-4mm} &
\includegraphics[width=0.108\textwidth]{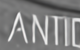} \hspace{-4mm}  
\\ 
\yulun{SRMDNF~\cite{zhang2018learning}} \hspace{-4mm} &
\yulun{D-DBPN~\cite{haris2018deep}}  \hspace{-4mm} &
\yulun{RDN (ours)} \hspace{-4mm}
\\
\end{tabular}
\end{adjustbox}
\\
\end{tabular}
\vspace{-3mm}
\caption{\yulun{Visual results on real-world images with scaling factor $\times4$.}}
\label{fig:result_SR_RGB_RealX4}
\vspace{-3mm}
\end{figure}

\begin{table*}[hbp]
\scriptsize
\center
\begin{center}
\caption{Quantitative results about gray-scale image denoising. Best and second best results are \textbf{highlighted} and \underline{underlined}}
\label{tab:results_psnr_denoise_gray}
\vspace{-3mm}
\begin{tabular}{|l|c|c|c|c|c|c|c|c|c|c|c|c|c|c|c|c|c|c|c|c|}
\hline
\multirow{2}{*}{Method} &  \multicolumn{4}{c|}{\ylzhang{Set12}} &  \multicolumn{4}{c|}{Kodak24} &  \multicolumn{4}{c|}{BSD68} &  \multicolumn{4}{c|}{Urban100}   
\\
\cline{2-17}
 & \ylzhang{10} & \ylzhang{30} & \ylzhang{50} & \ylzhang{70} & 10 & 30 & 50 & 70 & 10 & 30 & 50 & 70 & 10 & 30 & 50 & 70 
\\
\hline
\hline
BM3D~\cite{dabov2007image}
& \ylzhang{34.38}
 & \ylzhang{29.13}
  & \ylzhang{26.72}
   & \ylzhang{25.22}
   
& 34.39
 & 29.13
  & 26.99
   & 25.73
    & 33.31
     & 27.76
      & 25.62
       & 24.44
        & 34.47
         & 28.75
          & 25.94
           & 24.27
           
\\
\hline
TNRD~\cite{chen2017trainable}
& \ylzhang{34.27}
 & \ylzhang{28.63}
  & \ylzhang{26.81}
   & \ylzhang{24.12}
   
& 34.41
 & 28.87
  & 27.20
   & 24.95
    & 33.41
     & 27.66
      & 25.97
       & 23.83
        & 33.78
         & 27.49
          & 25.59
           & 22.67
           
\\
\hline
RED~\cite{mao2016image}
& \ylzhang{34.89}
 & \ylzhang{29.70}
  & \ylzhang{27.33}
   & \ylzhang{25.80}
   
& 35.02
 & 29.77
  & 27.66
   & 26.39
    & 33.99
     & 28.50
      & 26.37
       & 25.10
        & 34.91
         & 29.18
          & 26.51
           & 24.82
           
\\
\hline
DnCNN~\cite{zhang2017beyond}
& \ylzhang{34.78}
 & \ylzhang{29.53}
  & \ylzhang{27.18}
   & \ylzhang{25.52}
   
& 34.90
 & 29.62
  & 27.51
   & 26.08
    & 33.88
     & 28.36
      & 26.23
       & 24.90
        & 34.73
         & 28.88
          & 26.28
           & 24.36
           
\\
\hline
MemNet~\cite{tai2017memnet}
& \ylzhang{N/A}
 & \ylzhang{29.63}
  & \ylzhang{27.38}
   & \ylzhang{25.90}
   
& N/A
 & 29.72
  & 27.68
   & 26.42
    & N/A
     & 28.43
      & 26.35
       & 25.09
        & N/A
         & 29.10
          & 26.65
           & 25.01
           
\\
\hline
IRCNN~\cite{zhang2017learning}
& \ylzhang{34.72}
 & \ylzhang{29.45}
  & \ylzhang{27.14}
   & \ylzhang{N/A}
& 34.76
 & 29.53
  & 27.45
   & N/A
    & 33.74
     & 28.26
      & 26.15
       & N/A
        & 34.60
         & 28.85
          & 26.24
           & N/A
           
\\
\hline
\ylzhang{MWCNN}~\cite{liu2018multi}
& \ylzhang{N/A} & \ylzhang{N/A} & \ylzhang{\textbf{27.74}} & \ylzhang{N/A}
& \ylzhang{N/A} & \ylzhang{N/A} & \ylzhang{N/A} & \ylzhang{N/A} & \ylzhang{N/A} & \ylzhang{N/A} & \ylzhang{\textbf{26.53}}& \ylzhang{N/A} & \ylzhang{N/A} & \ylzhang{N/A} & \ylzhang{\underline{27.42}} & \ylzhang{N/A}           
\\
\hline
\ylzhang{N}$\ylzhang{^3}$\ylzhang{Net}~\cite{plotz2018neural}
& \ylzhang{N/A} & \ylzhang{N/A} & \ylzhang{27.43} & \ylzhang{25.90}
& \ylzhang{N/A} & \ylzhang{N/A} & \ylzhang{N/A} & \ylzhang{N/A} & \ylzhang{N/A} & \ylzhang{N/A} & \ylzhang{26.39} & \ylzhang{\textbf{25.14}} & \ylzhang{N/A} & \ylzhang{N/A} & \ylzhang{26.82} & \ylzhang{25.15}          
\\
\hline
\ylzhang{NLRN}~\cite{liu2018non}
& \ylzhang{N/A} & \ylzhang{N/A} & \ylzhang{27.64} & \ylzhang{N/A}
& \ylzhang{N/A} & \ylzhang{N/A} & \ylzhang{N/A} & \ylzhang{N/A} & \ylzhang{N/A} & \ylzhang{N/A} & \ylzhang{\underline{26.47}} & \ylzhang{N/A} & \ylzhang{N/A} & \ylzhang{29.94} & \ylzhang{27.38} & \ylzhang{\underline{25.66}}          
\\
\hline
FFDNet~\cite{zhang2017ffdnet}
& \ylzhang{34.65} & \ylzhang{29.61} & \ylzhang{27.32} & \ylzhang{25.81}
& 34.81
 & 29.70
  & 27.63
   & 26.34
    & 33.76
     & 28.39
      & 26.30
       & 25.04
        & 34.45
         & 29.03
          & 26.52
           & 24.86
           
\\
\hline
RDN (ours)
& \ylzhang{\underline{35.06}}
 & \ylzhang{\underline{29.94}}
  & \ylzhang{27.60}
   & \ylzhang{\underline{26.05}}
   
& \underline{35.17}
 & \underline{30.00}
  & \underline{27.85}
   & \underline{26.54}
    & \underline{34.00}
     & \underline{28.56}
      & {26.41}
       & {25.10}
        & \underline{35.41}
         & \underline{30.01}
          & {27.40}
           & {25.64}
           
\\
\hline
RDN+ (ours)
& \ylzhang{\textbf{35.08}}
 & \ylzhang{\textbf{29.97}}
  & \ylzhang{\underline{27.64}}
   & \ylzhang{\textbf{26.09}}
   
& \textbf{35.19}
 & \textbf{30.02}
  & \textbf{27.88}
   & \textbf{26.57}
    & \textbf{34.01}
     & \textbf{28.58}
      & {26.43}
       & \underline{25.12}
        & \textbf{35.45}
         & \textbf{30.08}
          & \textbf{27.47}
           & \textbf{25.71}
           
\\
\hline
           
\end{tabular}
\end{center}
\vspace{-5mm}
\end{table*}

\subsubsection{Super-Resolving Real-World Images}
\label{subsec:realworld}
\yulun{To further demonstrate the effectiveness of our proposed RDN, we super-resolve historical images with JPEG compression artifacts by $\times4$. We compare with two state-of-the-art methods: SRMDNF~\cite{zhang2018learning} and D-DBPN~\cite{haris2018deep}. As shown in Figure~\ref{fig:result_SR_RGB_RealX4}, the historical image contains letters ``HOME" and ``ANTI". Both SRMDNF and D-DBPN suffer from blurring and distorted artifacts. On the other hand, our RDN reconstruct sharper and more accurate results. These comparisons indicate the benefits of learning hierarchical features from the original input, allowing our RDN to perform robustly for different or unknown degradation models.}  


\begin{table*}[thbp]
\center
\begin{center}
\caption{Quantitative results about color image denoising. Best and second best results are \textbf{highlighted} and \underline{underlined}}
\label{tab:results_psnr_denoise_rgb}
\vspace{-3mm}
\begin{tabular}{|l|c|c|c|c|c|c|c|c|c|c|c|c|c|c|c|c|}
\hline
\multirow{2}{*}{Method} &  \multicolumn{4}{c|}{Kodak24} &  \multicolumn{4}{c|}{BSD68} &  \multicolumn{4}{c|}{Urban100}   
\\
\cline{2-13}
 & 10 & 30 & 50 & 70 & 10 & 30 & 50 & 70 & 10 & 30 & 50 & 70 
\\
\hline
\hline
CBM3D~\cite{dabov2007color}
& 36.57
 & 30.89
  & 28.63
   & 27.27
    & 35.91
     & 29.73
      & 27.38
       & 26.00
        & 36.00
         & 30.36
          & 27.94
           & 26.31
           
\\
\hline
TNRD~\cite{chen2017trainable}
& 34.33
 & 28.83
  & 27.17
   & 24.94
    & 33.36
     & 27.64
      & 25.96
       & 23.83
        & 33.60
         & 27.40
          & 25.52
           & 22.63
           
\\
\hline
RED~\cite{mao2016image}
& 34.91
 & 29.71
  & 27.62
   & 26.36
    & 33.89
     & 28.46
      & 26.35
       & 25.09
        & 34.59
         & 29.02
          & 26.40
           & 24.74
           
\\
\hline
DnCNN~\cite{zhang2017beyond}
& 36.98
 & 31.39
  & 29.16
   & 27.64
    & 36.31
     & 30.40
      & 28.01
       & 26.56
        & 36.21
         & 30.28
          & 28.16
           & 26.17
            
\\
\hline
MemNet~\cite{tai2017memnet}
& N/A
 & 29.67
  & 27.65
   & 26.40
    & N/A
     & 28.39
      & 26.33
       & 25.08
        & N/A
         & 28.93
          & 26.53
           & 24.93
           
\\
\hline
IRCNN~\cite{zhang2017learning}
& 36.70
 & 31.24
  & 28.93
   & N/A
    & 36.06
     & 30.22
      & 27.86
       & N/A
        & 35.81
         & 30.28
          & 27.69
           & N/A
           
\\
\hline
FFDNet~\cite{zhang2017ffdnet}
& 36.81
 & 31.39
  & 29.10
   & 27.68
    & 36.14
     & 30.31
      & 27.96
       & 26.53
        & 35.77
         & 30.53
          & 28.05
           & 26.39
           
\\
\hline
RDN (ours)
& \underline{37.31}
 & \underline{31.94}
  & \underline{29.66}
   & \underline{28.20}
    & \underline{36.47}
     & \underline{30.67}
      & \underline{28.31}
       & \underline{26.85}
        & \underline{36.69}
         & \underline{31.69}
          & \underline{29.29}
           & \underline{27.63}
           
\\
\hline
RDN+ (ours)
& \textbf{37.33}
 & \textbf{31.98}
  & \textbf{29.70}
   & \textbf{28.24}
    & \textbf{36.49}
     & \textbf{30.70}
      & \textbf{28.34}
       & \textbf{26.88}
        & \textbf{36.75}
         & \textbf{31.78}
          & \textbf{29.38}
           & \textbf{27.74}
           
\\
\hline
           
\end{tabular}
\end{center}
\end{table*}

\begin{figure*}[htpb]
\scriptsize
\centering
\begin{tabular}{cc}
\hspace{-0.4cm}
\begin{adjustbox}{valign=t}
\begin{tabular}{c}
\includegraphics[width=0.1955\textwidth]{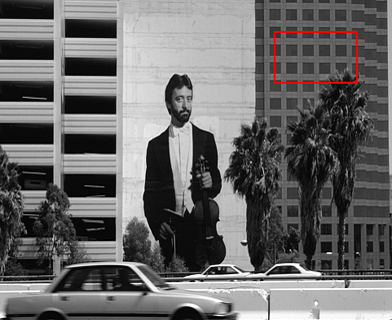}
\\
BSD68: 119082
\end{tabular}
\end{adjustbox}
\hspace{-0.46cm}
\begin{adjustbox}{valign=t}
\begin{tabular}{cccccc}
\includegraphics[width=0.138\textwidth]{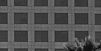} \hspace{-4mm} &
\includegraphics[width=0.138\textwidth]{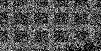} \hspace{-4mm} &
\includegraphics[width=0.138\textwidth]{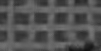} \hspace{-4mm} &
\includegraphics[width=0.138\textwidth]{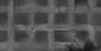} \hspace{-4mm} &
\includegraphics[width=0.138\textwidth]{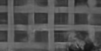} \hspace{-4mm} 
\\
GT \hspace{-4mm} &
Noisy ($\sigma$=50) \hspace{-4mm} &
BM3D~\cite{dabov2007image} \hspace{-4mm} &
TNRD~\cite{chen2017trainable} \hspace{-4mm} &
RED~\cite{mao2016image}  \hspace{-4mm}
\\
\includegraphics[width=0.138\textwidth]{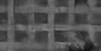} \hspace{-4mm} &
\includegraphics[width=0.138\textwidth]{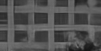} \hspace{-4mm} &
\includegraphics[width=0.138\textwidth]{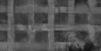} \hspace{-4mm} &
\includegraphics[width=0.138\textwidth]{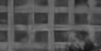} \hspace{-4mm} &
\includegraphics[width=0.138\textwidth]{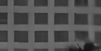} \hspace{-4mm}  
\\ 
DnCNN~\cite{zhang2017beyond} \hspace{-4mm} &
MemNet~\cite{tai2017memnet} \hspace{-4mm} &
IRCNN~\cite{zhang2017learning} \hspace{-4mm} &
FFDNet~\cite{zhang2017ffdnet}  \hspace{-4mm} &
RDN (ours) \hspace{-4mm}
\\
\end{tabular}
\end{adjustbox}
\vspace{2mm}
\\
\hspace{-0.4cm}
\begin{adjustbox}{valign=t}
\begin{tabular}{c}
\includegraphics[width=0.1955\textwidth]{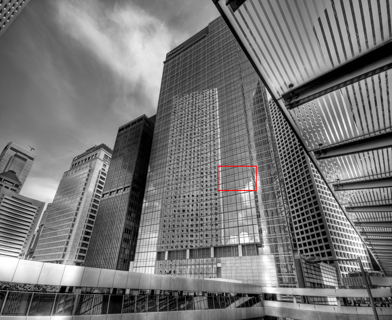}
\\
Urban100: img\_061
\end{tabular}
\end{adjustbox}
\hspace{-0.46cm}
\begin{adjustbox}{valign=t}
\begin{tabular}{cccccc}
\includegraphics[width=0.138\textwidth]{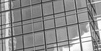} \hspace{-4mm} &
\includegraphics[width=0.138\textwidth]{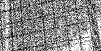} \hspace{-4mm} &
\includegraphics[width=0.138\textwidth]{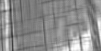} \hspace{-4mm} &
\includegraphics[width=0.138\textwidth]{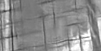} \hspace{-4mm} &
\includegraphics[width=0.138\textwidth]{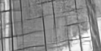} \hspace{-4mm} 
\\
GT \hspace{-4mm} &
Noisy ($\sigma$=50) \hspace{-4mm} &
BM3D~\cite{dabov2007image} \hspace{-4mm} &
TNRD~\cite{chen2017trainable} \hspace{-4mm} &
RED~\cite{mao2016image}  \hspace{-4mm}
\\
\includegraphics[width=0.138\textwidth]{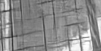} \hspace{-4mm} &
\includegraphics[width=0.138\textwidth]{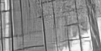} \hspace{-4mm} &
\includegraphics[width=0.138\textwidth]{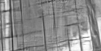} \hspace{-4mm} &
\includegraphics[width=0.138\textwidth]{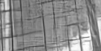} \hspace{-4mm} &
\includegraphics[width=0.138\textwidth]{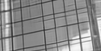} \hspace{-4mm}  
\\ 
DnCNN~\cite{zhang2017beyond} \hspace{-4mm} &
MemNet~\cite{tai2017memnet} \hspace{-4mm} &
IRCNN~\cite{zhang2017learning} \hspace{-4mm} &
FFDNet~\cite{zhang2017ffdnet}  \hspace{-4mm} &
RDN (ours) \hspace{-4mm}
\\
\end{tabular}
\end{adjustbox}
\end{tabular}
\vspace{-3mm}
\caption{Gray-scale image denoising results with noise level $\sigma$ = 50.}
\label{fig:result_DN_Gray_N50}
\vspace{-3mm}
\end{figure*}

\vspace{-3mm}
\subsection{Image Denoising}
\label{subsec:denoising}
We compare our RDN with recently leading Gaussian denoising methods: BM3D~\cite{dabov2007image}, CBM3D~\cite{dabov2007color}, TNRD~\cite{chen2017trainable}, RED~\cite{mao2016image}, DnCNN~\cite{zhang2017beyond}, MemNet~\cite{tai2017memnet}, IRCNN~\cite{zhang2017learning}, \ylzhang{MWCNN}~\cite{liu2018multi}, \ylzhang{N}$\ylzhang{^3}$\ylzhang{Net}~\cite{plotz2018neural}, \ylzhang{NLRN}~\cite{liu2018non}, and FFDNet~\cite{zhang2017ffdnet}. Kodak24~\footnote{\href{http://r0k.us/graphics/kodak/}{http://r0k.us/graphics/kodak/}}, BSD68~\cite{martin2001database}, and Urban100~\cite{huang2015single} are used for gray-scale and color image denoising. \ylzhang{Set12}~\cite{zhang2017beyond} \ylzhang{is also used to test gray-scale image denoising.} Noisy images are obtained by adding AWGN noises of different levels to clean images.

\vspace{-2mm}
\subsubsection{Gray-scale Image Denoising}
The PSNR results are shown in Table~\ref{tab:results_psnr_denoise_gray}. One can see that on all the \ylzhang{4} test sets with 4 noise levels, our RDN+ achieves \ylzhang{higher average PSNR values than most of compared methods}. On average, for noise level $\sigma=50$, our RDN achieves \ylzhang{0.28 dB}, 0.22 dB, 0.11 dB, and 0.88 dB gains over FFDNet~\cite{zhang2017ffdnet} on \ylzhang{four} test sets respectively. Gains on Urban100 become larger, which is mainly because our method takes advantage of a larger scope of context information with hierarchical features. Moreover, for noise levels $\sigma$ = 30, 50, and 70, the gains over BM3D of RDN are larger than 0.7 dB, breaking the estimated PSNR bound (0.7 dB) over BM3D in~\cite{levin2012patch}. It should also be noted that our RDN achieves moderate gains over MemNet on BSD68. This is mainly because BSD68 is formed from 100 validation images in Berkeley Segmentation Dataset (BSD)~\cite{martin2001database}. But MemNet~\cite{tai2017memnet} used these 100 validation images and 200 training images in BSD as training data. So it's reasonable for MemNet to perform pretty well on BSD68. \ylzhang{Moreover,} \ylzhang{MWCNN}~\cite{liu2018multi}, \ylzhang{N}$\ylzhang{^3}$\ylzhang{Net}~\cite{plotz2018neural}, \ylzhang{and NLRN}~\cite{liu2018non} \ylzhang{also have achieved good performance for some noise levels, which indicates that Wavelet transformation and non-local processing are promissing candidates for further image denoising improvements.} 

We show visual gray-scale denoised results of different methods in Figure~\ref{fig:result_DN_Gray_N50}. We can see that BM3D preserves image structure to some degree, but fails to remove noise deeply. TNRD~\cite{chen2017trainable} tends to generate some artifacts in the smooth region. RED~\cite{mao2016image}, DnCNN~\cite{zhang2017beyond}, MemNet~\cite{tai2017memnet}, and IRCNN~\cite{zhang2017learning} would over-smooth edges. The main reason should be the limited network ability for high noise level (\eg, $\sigma$ = 50). In contrast, our RDN can remove noise greatly and recover more details (\eg, the tiny lines in ``img\_061''). Also, the gray-scale visual results by our RDN in the smooth region are more faithful to the clean images (\eg, smooth regions in ``119082'' and ``img\_061''). 

\begin{figure*}[htpb]
\scriptsize
\centering
\begin{tabular}{cc}
\hspace{-0.4cm}
\begin{adjustbox}{valign=t}
\begin{tabular}{c}
\includegraphics[width=0.1955\textwidth]{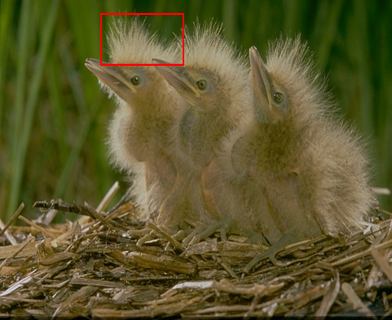}
\\
BSD68: 163085
\end{tabular}
\end{adjustbox}
\hspace{-0.46cm}
\begin{adjustbox}{valign=t}
\begin{tabular}{cccccc}
\includegraphics[width=0.138\textwidth]{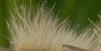} \hspace{-4mm} &
\includegraphics[width=0.138\textwidth]{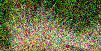} \hspace{-4mm} &
\includegraphics[width=0.138\textwidth]{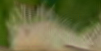} \hspace{-4mm} &
\includegraphics[width=0.138\textwidth]{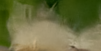} \hspace{-4mm} &
\includegraphics[width=0.138\textwidth]{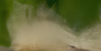} \hspace{-4mm} 
\\
GT \hspace{-4mm} &
Noisy ($\sigma$=50) \hspace{-4mm} &
CBM3D~\cite{dabov2007color} \hspace{-4mm} &
TNRD~\cite{chen2017trainable} \hspace{-4mm} &
RED~\cite{mao2016image}  \hspace{-4mm}
\\
\includegraphics[width=0.138\textwidth]{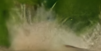} \hspace{-4mm} &
\includegraphics[width=0.138\textwidth]{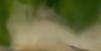} \hspace{-4mm} &
\includegraphics[width=0.138\textwidth]{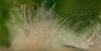} \hspace{-4mm} &
\includegraphics[width=0.138\textwidth]{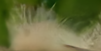} \hspace{-4mm} &
\includegraphics[width=0.138\textwidth]{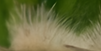} \hspace{-4mm}  
\\ 
DnCNN~\cite{zhang2017beyond} \hspace{-4mm} &
MemNet~\cite{tai2017memnet} \hspace{-4mm} &
IRCNN~\cite{zhang2017learning} \hspace{-4mm} &
FFDNet~\cite{zhang2017ffdnet}  \hspace{-4mm} &
RDN (ours) \hspace{-4mm}
\\
\end{tabular}
\end{adjustbox}
\vspace{2mm}
\\
\hspace{-0.4cm}
\begin{adjustbox}{valign=t}
\begin{tabular}{c}
\includegraphics[width=0.1955\textwidth]{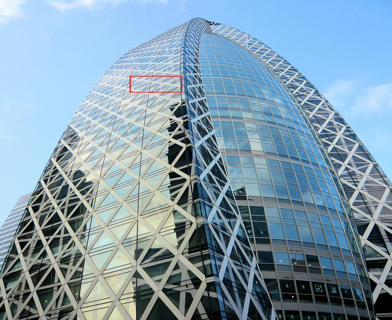}
\\
Urban100: img\_039
\end{tabular}
\end{adjustbox}
\hspace{-0.46cm}
\begin{adjustbox}{valign=t}
\begin{tabular}{cccccc}
\includegraphics[width=0.138\textwidth]{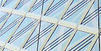} \hspace{-4mm} &
\includegraphics[width=0.138\textwidth]{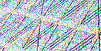} \hspace{-4mm} &
\includegraphics[width=0.138\textwidth]{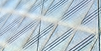} \hspace{-4mm} &
\includegraphics[width=0.138\textwidth]{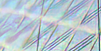} \hspace{-4mm} &
\includegraphics[width=0.138\textwidth]{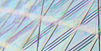} \hspace{-4mm} 
\\
GT \hspace{-4mm} &
Noisy ($\sigma$=50) \hspace{-4mm} &
CBM3D~\cite{dabov2007color} \hspace{-4mm} &
TNRD~\cite{chen2017trainable} \hspace{-4mm} &
RED~\cite{mao2016image}  \hspace{-4mm}
\\
\includegraphics[width=0.138\textwidth]{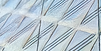} \hspace{-4mm} &
\includegraphics[width=0.138\textwidth]{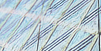} \hspace{-4mm} &
\includegraphics[width=0.138\textwidth]{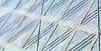} \hspace{-4mm} &
\includegraphics[width=0.138\textwidth]{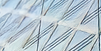} \hspace{-4mm} &
\includegraphics[width=0.138\textwidth]{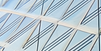} \hspace{-4mm}  
\\ 
DnCNN~\cite{zhang2017beyond} \hspace{-4mm} &
MemNet~\cite{tai2017memnet} \hspace{-4mm} &
IRCNN~\cite{zhang2017learning} \hspace{-4mm} &
FFDNet~\cite{zhang2017ffdnet}  \hspace{-4mm} &
RDN (ours) \hspace{-4mm}
\\
\end{tabular}
\end{adjustbox}
\end{tabular}
\vspace{-2mm}
\caption{Color image denoising results with noise level $\sigma$ = 50.}
\label{fig:result_DN_RGB_N50}
\vspace{-3mm}
\end{figure*}

\begin{table*}[htbp]
\center
\begin{center}
\caption{Quantitative results about image compression artifact reduction. Best and second best results are \underline{highlighted} and {underlined}}
\label{tab:results_psnr_ssim_car_y}
\vspace{-3mm}
\begin{tabular*}{178.2mm}{|c|c|c@{\extracolsep{-0.99mm}}|c|c|c|c|c|c|c|c|c|c|c|c|c|c|c|c|c|c|c|}
\hline
\multirow{2}{*}{Dataset} & \multirow{2}{*}{Quality} &  \multicolumn{2}{c|}{JPEG} &  \multicolumn{2}{c|}{SA-DCT~\cite{foi2007pointwise}} &  \multicolumn{2}{c|}{ARCNN~\cite{dong2015compression}} &  \multicolumn{2}{c|}{TNRD~\cite{chen2017trainable}} &  \multicolumn{2}{c|}{DnCNN~\cite{zhang2017beyond}} &  \multicolumn{2}{c|}{RDN (ours)} &  \multicolumn{2}{c|}{RDN+ (ours)}   
\\
\cline{3-16}
&  & PSNR & SSIM & PSNR & SSIM & PSNR & SSIM & PSNR & SSIM & PSNR & SSIM & PSNR & SSIM & PSNR & SSIM 
\\
\hline
\hline
\multirow{4}{*}{LIVE1} & 10 
& 27.77
 & 0.7905
  & 28.65
   & 0.8093
    & 28.98
     & 0.8217
      & 29.15
       & 0.8111
        & {29.19}
         & {0.8123}
          & \underline{29.67}
           & \underline{0.8247} 
            & \textbf{29.70}
             & \textbf{0.8252}

\\
& 20 
& 30.07
 & 0.8683
  & 30.81
   & 0.8781
    & 31.29
     & 0.8871
      & 31.46
       & 0.8769
        & {31.59}
         & {0.8802}
          & \underline{32.07}
           & \underline{0.8882}
            & \textbf{32.10}
             & \textbf{0.8886}

\\
& 30 
& 31.41
 & 0.9000
  & 32.08
   & 0.9078
    & 32.69
     & 0.9166
      & 32.84
       & 0.9059
        & {32.98}
         & {0.9090}
          & \underline{33.51}
           & \underline{0.9153}
            & \textbf{33.54}
             & \textbf{0.9156}

\\
& 40 
& 32.35
 & 0.9173
  & 32.99
   & 0.9240
    & 33.63
     & 0.9306
      & N/A
       & N/A
        & {33.96}
         & {0.9247}
          & \underline{34.51}
           & \underline{0.9302}
            & \textbf{34.54}
             & \textbf{0.9304}

\\
\hline 
\hline
\multirow{4}{*}{Classic5} & 10 
& 27.82
 & 0.7800
  & 28.88
   & 0.8071
    & 29.04
     & 0.8111
      & 29.28
       & 0.7992
        & {29.40}
         & {0.8026}
          & \underline{30.00}
           & \underline{0.8188}
            & \textbf{30.03}
             & \textbf{0.8194}

\\
& 20 
& 30.12
 & 0.8541
  & 30.92
   & 0.8663
    & 31.16
     & 0.8694
      & 31.47
       & 0.8576
        & {31.63}
         & {0.8610}
          & \underline{32.15}
           & \underline{0.8699}
            & \textbf{32.19}
             & \textbf{0.8704}

\\
& 30 
& 31.48
 & 0.8844
  & 32.14
   & 0.8914
    & 32.52
     & 0.8967
      & 32.78
       & 0.8837
        & {32.91}
         & {0.8861}
          & \underline{33.43}
           & \underline{0.8930}
            & \textbf{33.46}
             & \textbf{0.8932}

\\
& 40 
& 32.43
 & 0.9011
  & 33.00
   & 0.9055
    & 33.34
     & 0.9101
      & N/A
       & N/A
        & {33.77}
         & {0.9003}
          & \underline{34.27}
           & \underline{0.9061}
            & \textbf{34.29}
             & \textbf{0.9063}

\\
\hline          
\end{tabular*}
\end{center}
\vspace{-2mm}
\end{table*}

\begin{figure*}[htpb]
\scriptsize
\centering
\begin{tabular}{cc}
\hspace{-0.338cm}
\begin{adjustbox}{valign=t}
\begin{tabular}{ccccccc}
\includegraphics[width=0.136\textwidth]{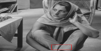} \hspace{-4.2mm} &
\includegraphics[width=0.136\textwidth]{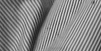} \hspace{-4.2mm} &
\includegraphics[width=0.136\textwidth]{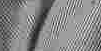} \hspace{-4.2mm} &
\includegraphics[width=0.136\textwidth]{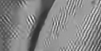} \hspace{-4.2mm} &
\includegraphics[width=0.136\textwidth]{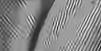} \hspace{-4.2mm} &
\includegraphics[width=0.136\textwidth]{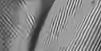} \hspace{-4.2mm} &
\includegraphics[width=0.136\textwidth]{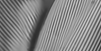} \hspace{-4.2mm} 
\\

\includegraphics[width=0.136\textwidth]{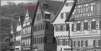} \hspace{-4.2mm} &
\includegraphics[width=0.136\textwidth]{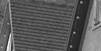} \hspace{-4.2mm} &
\includegraphics[width=0.136\textwidth]{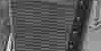} \hspace{-4.2mm} &
\includegraphics[width=0.136\textwidth]{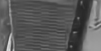} \hspace{-4.2mm} &
\includegraphics[width=0.136\textwidth]{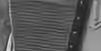} \hspace{-4.2mm} &
\includegraphics[width=0.136\textwidth]{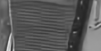} \hspace{-4.2mm} &
\includegraphics[width=0.136\textwidth]{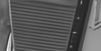} \hspace{-4.2mm} 
\\

\includegraphics[width=0.136\textwidth]{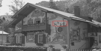} \hspace{-4.2mm} &
\includegraphics[width=0.136\textwidth]{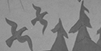} \hspace{-4.2mm} &
\includegraphics[width=0.136\textwidth]{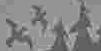} \hspace{-4.2mm} &
\includegraphics[width=0.136\textwidth]{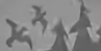} \hspace{-4.2mm} &
\includegraphics[width=0.136\textwidth]{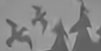} \hspace{-4.2mm} &
\includegraphics[width=0.136\textwidth]{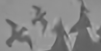} \hspace{-4.2mm} &
\includegraphics[width=0.136\textwidth]{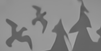} \hspace{-4.2mm} 
\\

\includegraphics[width=0.136\textwidth]{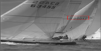} \hspace{-4.2mm} &
\includegraphics[width=0.136\textwidth]{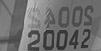} \hspace{-4.2mm} &
\includegraphics[width=0.136\textwidth]{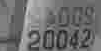} \hspace{-4.2mm} &
\includegraphics[width=0.136\textwidth]{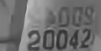} \hspace{-4.2mm} &
\includegraphics[width=0.136\textwidth]{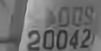} \hspace{-4.2mm} &
\includegraphics[width=0.136\textwidth]{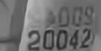} \hspace{-4.2mm} &
\includegraphics[width=0.136\textwidth]{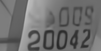} \hspace{-4.2mm} 
\\
 \hspace{-4.2mm} &
GT \hspace{-4.2mm} &
JPEG ($q$=10) \hspace{-4.2mm} &
ARCNN~\cite{dong2015compression} \hspace{-4.2mm} &
TNRD~\cite{chen2017trainable} \hspace{-4.2mm} &
DnCNN~\cite{zhang2017beyond} \hspace{-4.2mm} &
RDN (ours) \hspace{-4.2mm}
\\
\end{tabular}
\end{adjustbox}

\end{tabular}
\vspace{-3mm}
\caption{Image compression artifacts reduction results with JPEG quality $q$ = 10.}
\label{fig:result_CAR_Y_Q10}
\vspace{-2mm}
\end{figure*}

\vspace{-2mm}
\subsubsection{Color Image Denoising}
We generate noisy color images by adding AWGN noise to clean RGB images with different noise levels $\sigma$ = 10, 30, 50, and 70. The PSNR results are listed in Table~\ref{tab:results_psnr_denoise_rgb}. We apply gray image denoising methods (\eg, MemNet~\cite{tai2017memnet}) for color image denoising channel by channel. Lager gains over MemNet~\cite{tai2017memnet} of our RDN indicate that denoising color images jointly perform better than denoising each channel separately. Take $\sigma$=50 as an example, our RDN obtains 0.56 dB, 0.35, and 1.24 dB improvements over FFDNet~\cite{zhang2017ffdnet} on three test sets respectively. Residual learning and dense feature fusion allows RDN to go wider and deeper, obtain hierarchical features, and achieve better performance. 

We also show color image denoising visual results in Figure~\ref{fig:result_DN_RGB_N50}. CBM3D~\cite{dabov2007color} tends to produce artifacts along the edges. TNRD~\cite{chen2017trainable} produces artifacts in the smooth area and is unable to recover clear edges. RED~\cite{mao2016image}, DnCNN~\cite{zhang2017beyond}, MemNet~\cite{tai2017memnet}, IRCNN~\cite{zhang2017learning}, and FFDNet~\cite{zhang2017ffdnet} could produce blurring artifacts along edges (\eg, the structural lines in ``img\_039"). Because RED~\cite{mao2016image} and MemNet~\cite{tai2017memnet} were designed for gray image denoising. In our experiments on color image denoising, we conduct RED~\cite{mao2016image} and MemNet~\cite{tai2017memnet} in each channel. Although DnCNN~\cite{zhang2017beyond}, IRCNN~\cite{zhang2017learning}, and FFDNet~\cite{zhang2017ffdnet} directly denoise noisy color images in three channels, they either fail to recover sharp edges and clean smooth area. In contrast, our RDN can recover shaper edges and cleaner smooth area.

\vspace{-5mm}
\yulun{\subsubsection{Real-World Color Image Denoising}}
\yulun{To further demonstrate the effectiveness of our proposed RDN, we compare it with recent leading method MCWNNM~\cite{xu2017multi} on real noisy image. Following the same settings as~\cite{xu2017multi}, we estimate the noise levels $\left ( \sigma _{r},\sigma _{g},\sigma _{b} \right )$ via some noise estimation methods~\cite{chen2015efficient}. Inspired by~\cite{xu2017multi}, we finetune RDN with the noise level $\sigma$=$\sqrt{\left ( \sigma _{r}^{2} + \sigma _{g}^{2} + \sigma _{b}^{2} \right )/3}$. Due to limited space, we only show the denoised results on the real noisy image ``Dog" (with 192$\times$192 pixels)~\cite{lebrun2015noise}, which doesn't have ground truth.}

\begin{figure}[t]
\scriptsize
\centering
\begin{tabular}{cc}
\hspace{-0.4cm}
\begin{adjustbox}{valign=t}
\begin{tabular}{c}
\includegraphics[width=0.1420\textwidth]{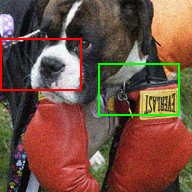}
\\
\end{tabular}
\end{adjustbox}
\hspace{-0.46cm}
\begin{adjustbox}{valign=t}
\begin{tabular}{cccccc}
\includegraphics[width=0.108\textwidth]{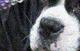} \hspace{-4mm} &
\includegraphics[width=0.108\textwidth]{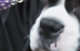} \hspace{-4mm} &
\includegraphics[width=0.108\textwidth]{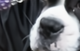} \hspace{-4mm} 
\\
\includegraphics[width=0.108\textwidth]{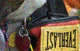} \hspace{-4mm} &
\includegraphics[width=0.108\textwidth]{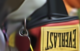} \hspace{-4mm} &
\includegraphics[width=0.108\textwidth]{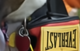} \hspace{-4mm}  
\\ 
\yulun{Noisy} \hspace{-4mm} &
\yulun{MCWNNM~\cite{xu2017multi}}  \hspace{-4mm} &
\yulun{RDN (ours)} \hspace{-4mm}
\\
\end{tabular}
\end{adjustbox}
\end{tabular}
\vspace{-3mm}
\caption{\yulun{Visual denoised results of the real noisy image ``Dog". The noise level of R, G, B channels are estimated as 16.8, 17.0, and 16.6 respectively.}}
\label{fig:result_DN_RGB_Real}
\vspace{-5mm}
\end{figure}

\yulun{We show real-world color image denoising in Fig.~\ref{fig:result_DN_RGB_Real}. Although MCWNNM considers the specific noise levels of each channel, its result still suffers from over-smoothing artifacts and loses some details (\eg, the moustache). In contrast, our RDN handles the noise better and preserves more details and shaper edges, which indicates that RDN is also suitable for real applications.}

\begin{figure*}[t]
\scriptsize
\centering
\begin{tabular}{cc}
\hspace{-0.4cm}
\begin{adjustbox}{valign=t}
\begin{tabular}{c}
\includegraphics[width=0.220\textwidth,height=0.1726\textwidth]{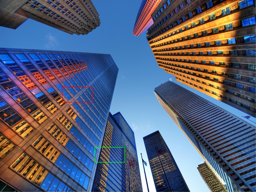}
\\
\yulun{Urban100: img\_012}
\end{tabular}
\end{adjustbox}
\hspace{-0.46cm}
\begin{adjustbox}{valign=t}
\begin{tabular}{cccccc}
\includegraphics[width=0.168\textwidth]{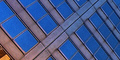} \hspace{-4mm} &
\includegraphics[width=0.168\textwidth]{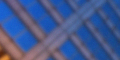} \hspace{-4mm} &
\includegraphics[width=0.168\textwidth]{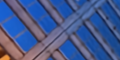} \hspace{-4mm} &
\includegraphics[width=0.168\textwidth]{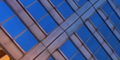} \hspace{-4mm} 
\\
\includegraphics[width=0.168\textwidth]{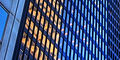} \hspace{-4mm} &
\includegraphics[width=0.168\textwidth]{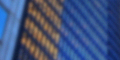} \hspace{-4mm} &
\includegraphics[width=0.168\textwidth]{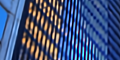} \hspace{-4mm} &
\includegraphics[width=0.168\textwidth]{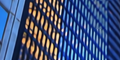} \hspace{-4mm}  
\\ 
\yulun{HQ} \hspace{-4mm} &
\yulun{Blurry} \hspace{-4mm} &
\yulun{IRCNN~\cite{zhang2017learning}}  \hspace{-4mm} &
\yulun{RDN (ours)} \hspace{-4mm}
\end{tabular}
\end{adjustbox}
\end{tabular}
\vspace{-3mm}
\caption{\yulun{Visual results on image deblurring.}}
\label{fig:result_deblur_RGB}
\vspace{-2mm}
\end{figure*}

\begin{figure*}[t]
\scriptsize
\centering
\begin{tabular}{cc}
\hspace{-0.4cm}
\begin{adjustbox}{valign=t}
\begin{tabular}{c}
\includegraphics[width=0.1955\textwidth]{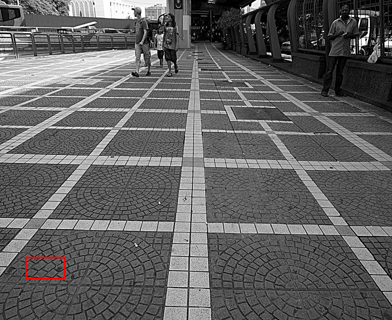}
\\
Urban100: img\_095 ($\times$8)
\end{tabular}
\end{adjustbox}
\hspace{-0.46cm}
\begin{adjustbox}{valign=t}
\begin{tabular}{cccccc}
\includegraphics[width=0.138\textwidth]{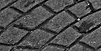} \hspace{-4mm} &
\includegraphics[width=0.138\textwidth]{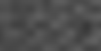} \hspace{-4mm} &
\includegraphics[width=0.138\textwidth]{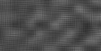} \hspace{-4mm} &
\includegraphics[width=0.138\textwidth]{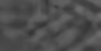} \hspace{-4mm} &
\includegraphics[width=0.138\textwidth]{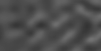} \hspace{-4mm} 
\\
GT \hspace{-4mm} &
Bicubic \hspace{-4mm} &
SRCNN~\cite{dong2016image} \hspace{-4mm} &
SCN~\cite{wang2015deep} \hspace{-4mm} &
VDSR~\cite{kim2016accurate}
\\
\includegraphics[width=0.138\textwidth]{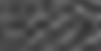} \hspace{-4mm} &
\includegraphics[width=0.138\textwidth]{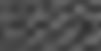} \hspace{-4mm} &
\includegraphics[width=0.138\textwidth]{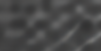} \hspace{-4mm} &
\includegraphics[width=0.138\textwidth]{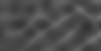} \hspace{-4mm} &
\includegraphics[width=0.138\textwidth]{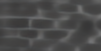} \hspace{-4mm}  
\\ 
MemNet~\cite{tai2017memnet} \hspace{-4mm} &
MSLapSRN~\cite{lai2018MSLapSRN} \hspace{-4mm} &
EDSR~\cite{lim2017enhanced} \hspace{-4mm} &
D-DBPN~\cite{haris2018deep}  \hspace{-4mm} &
RDN (ours) \hspace{-4.2mm}
\\
\\
\end{tabular}
\end{adjustbox}
\end{tabular}
\vspace{-3mm}
\caption{Failure cases for image super-resolution ($\times8$).}
\label{fig:result_SR_BIX8_failure}
\vspace{-3mm}
\end{figure*}

\vspace{3mm}
\subsection{Image Compression Artifact Reduction}
We further apply our RDN to reduce image compression artifacts. We compare our RDN with SA-DCT~\cite{foi2007pointwise}, ARCNN~\cite{dong2015compression}, TNRD~\cite{chen2017trainable}, and DnCNN~\cite{zhang2017beyond}. We use Matlab JPEG encoder~\cite{jancsary2012loss} to generate compressed test images from LIVE1~\cite{sheikh2005live} and Classic5~\cite{foi2007pointwise}. Four JPEG quality settings $q$ = 10, 20, 30, 40 are used in Matlab JPEG encoder. Here, we only focus on the compression artifact reduction (CAR) of Y channel (in YCbCr space) to keep fair comparison with other methods.

We report PSNR/SSIM values in Table~\ref{tab:results_psnr_ssim_car_y}. As we can see, our RDN and RDN+ achieve higher PSNR and SSIM values on LIVE1 and Classic5 with all JPEG qualities than other compared methods. Taking $q=10$ as an example, our RDN achieves 0.48 dB and 0.60 dB improvements over DnCNN~\cite{zhang2017beyond} in terms of PSNR. Even in such a challenging case (very low compression quality), our RDN can still obtain great performance gains over others. Similar improvements are also significant for other compression qualities. These comparisons further demonstrate the effectiveness of our proposed RDN.

Visual comparisons are further shown in Figure~\ref{fig:result_CAR_Y_Q10}, where we provide comparisons under very low image quality ($q$=10). Although ARCNN~\cite{dong2015compression}, TNRD~\cite{chen2017trainable}, and DnCNN~\cite{zhang2017beyond} can remove blocking artifacts to some degree, they also over-smooth some details (\eg, 1st and 2nd rows in Figure~\ref{fig:result_CAR_Y_Q10}) and cannot deeply remove the compression artifacts around content structures (\eg, 3rd and 4th rows in Figure~\ref{fig:result_CAR_Y_Q10}). While, RDN has stronger network representation ability to distinguish compression artifacts and content information better. As a result, RDN recovers more details with consistent content structures.  

\begin{table}[t]
\scriptsize
\centering
\begin{center}
\caption{\yulun{PSNR (dB)/SSIM results about image deblurring.}}
\label{tab:results_psnr_ssim_deblur}
\vspace{-3mm}
\begin{tabular}{|c|c|c|c|c|c|c|c|c|c|c|c|c|c|c|c|c|}
\hline
\multirow{2}{*}{\yulun{Set}} &  \multicolumn{2}{c|}{\yulun{McMaster18}} &  \multicolumn{2}{c|}{\yulun{Kodak24}} & \multicolumn{2}{c|}{\yulun{Urban100}}
\\
\cline{2-7}
 & \yulun{PSNR} & \yulun{SSIM} & \yulun{PSNR} & \yulun{SSIM} & \yulun{PSNR} & \yulun{SSIM}
\\
\hline

\yulun{Blurry} & \yulun{27.00} & \yulun{0.7817} & \yulun{26.09} & \yulun{0.7142} & \yulun{22.38} & \yulun{0.6732}
\\
\hline
\yulun{IRCNN~\cite{zhang2017learning}}   & \yulun{32.50} & \yulun{0.8961} & \yulun{30.40} & \yulun{0.8513} & \yulun{27.70} & \yulun{0.8577}                              
\\
\hline
\yulun{RDN (ours)}     & \yulun{33.60} & \yulun{0.9157} & \yulun{30.88} & \yulun{0.8718} & \yulun{29.34} & \yulun{0.8886}                              
\\
\hline          
\end{tabular}
\end{center}
\vspace{-4mm}
\end{table}

\vspace{-6mm}
\yulun{\subsection{Image Deblurring}}
\yulun{A common practice to synthesize blurry images is first apply a blur kernel and then add Gausian noise with noise level $\sigma$~\cite{zhang2017learning,dong2019denoising}. There are several blur kernels, such as Gaussian and motion blur kernels. Here, we focus on the commonly-used 25$\times$25 Gaussian blur kernel of standard deviation 1.6. The additive Gaussian noise ($\sigma=2$) is added to the blurry images.}

\yulun{We compare with IRCNN~\cite{zhang2017learning} and use McMaster18~\cite{zhang2017learning}, Kodak24, and Urban100 as test sets. We provide quantitative results in Table~\ref{tab:results_psnr_ssim_deblur}, where our RDN achieves large improvements over IRCNN for each test set. We further show visual results in Figure~\ref{fig:result_deblur_RGB}, where IRCNN still outputs some blurry structures. While, our RDN reconstructs much sharper edges and tiny details. These quantitative and qualitative comparisons demonstrate that our RDN also performs well for image deblurring.}

\vspace{2mm}
\section{Discussions}
Here, we give a brief view of the benefits and limitations of our RDN and challenges in image restoration.

\textbf{Benefits of RDN}. \yulun{RDB serves as the basic build module in RDN, which takes advantage of local and global feature fusion, obtaining very powerful representational ability. RDN uses less network parameters than residual network while achieves better performance than dense network, leading to a good tradeoff between model size and performance. RDN can be directly applied or generilized to several image restoration tasks with promissing performance.} 



\textbf{Limitations of RDN}. \yulun{In some challenging cases (\eg, large scaling factor), RDN may fail to obtain proper details. As shown in Figure~\ref{fig:result_SR_BIX8_failure}, although other methods fail to recover proper structures, our RDN cannot generates right structures either. The main reasons of this failure case may be that our RDN cannot recover similar textures based on the limited input information. Instead, RDN would generate most likely texture patterns learned from the training data.}


\textbf{Challenges in Image Restoration}. Extreme cases make the image restoration tasks much harder, such as very large scaling factors for image SR, heavy noise for image DN, low JPEG quality in image CAR, \yulun{and heavy blurring artifacts in image deblurring}. Complex desegregation processes in the real world make it difficult for us to formulate the degradation process. Then it may make the data preparation and network training harder.   

\yulun{\textbf{Future Works}. We believe that RDN can be further applied to other image restoration tasks, such as image demosaicing, derain, and dehazing. On the other hand, it's worth to further improve the performance of RDN. As indicated in Figure~\ref{fig:result_SR_BIX8_failure}, RDN generates better local structures than others, while the global recovered structures are wrong. How to learn stronger local and global feature representations is worth to investigate. Plus, RDN may further benefit from adversarial training, which may help to alleviate some blurring and over-smoothing artifacts.}

\section{Conclusions}
\yulun{In this work, based on our proposed residual dense block (RDB), we propose deep residual dense network (RDN) for image restoration. RDB allows direct connections from preceding RDB to each Conv layer of current RDB, which results in a contiguous memory (CM) mechanism. We proposed LFF to adaptively preserve the information from the current and previous RDBs. With the usage of LRL, the flow of gradient and information can be further improved and training wider network becomes more stable. Extensive benchmark and real-world evaluations well demonstrate that our RDN achieves superiority over state-of-the-art methods for several image restoration tasks.}



\ifCLASSOPTIONcompsoc
  \section*{Acknowledgments}
\else
  \section*{Acknowledgment}
\fi

This research is supported in part by the NSF IIS award 1651902, ONR Young Investigator Award N00014-14-1-0484, and U.S. Army Research Office Award W911NF-17-1-0367.

\ifCLASSOPTIONcaptionsoff
  \newpage
\fi



\vspace{-3mm}
\bibliographystyle{IEEEtran}
\bibliography{SR_journal_bib}
%

%
\balance
\begin{IEEEbiography}[{\includegraphics[width=1in,height=1.25in,clip,keepaspectratio]{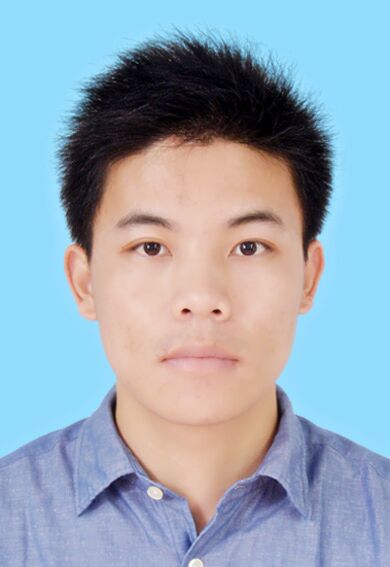}}]{Yulun Zhang} received B.E. degree from School of Electronic Engineering, Xidian University, China, in 2013 and M.E. degree from Department of Automation, Tsinghua University, China, in 2017. He is currently pursuing the Ph.D. degree with the Department of ECE, Northeastern University, USA. He was the receipt of the Best Student Paper Award at IEEE International Conference on Visual Communication and Image Processing(VCIP) in 2015. He also won the Best Paper Award at IEEE International Conference on Computer Vision (ICCV) RLQ Workshop in 2019. His research interests include image restoration and deep learning.
\vspace{-14mm}
\end{IEEEbiography}

\begin{IEEEbiography}[{\includegraphics[width=1in,height=1.25in,clip,keepaspectratio]{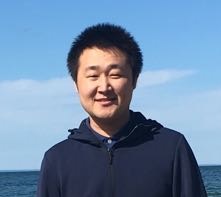}}]{Yapeng Tian} received the B.E. degree in electronic engineering from Xidian University, Xi’an, China, in 2013, M.E. degree in electronic engineering at Tsinghua University, Beijing, China, in 2017, and is currently working toward the PhD degree in department of computer science at University of Rochester, USA. His research interests include audio-visual scene understanding and low-level vision. 
\vspace{-15mm}
\end{IEEEbiography}

\begin{IEEEbiography}[{\includegraphics[width=1in,height=1.25in,clip,keepaspectratio]{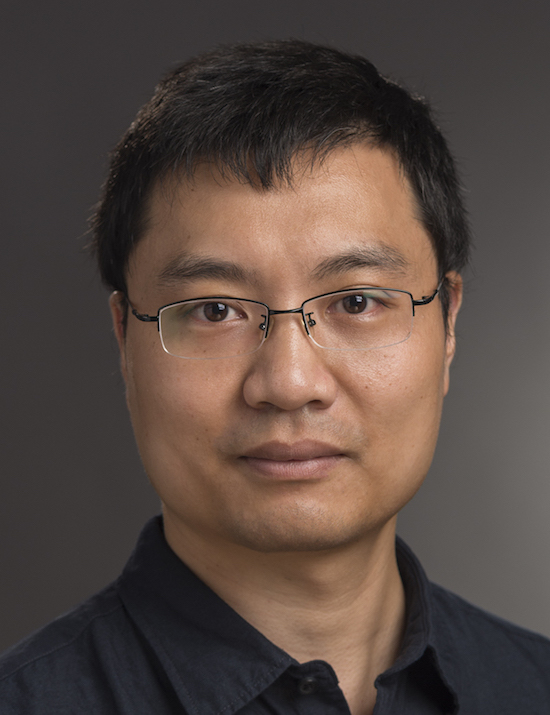}}]{Yu Kong} received B.Eng. degree in automation from Anhui University in 2006, and PhD degree in computer science from Beijing Institute of Technology, China, in 2012. He is now a tenure-track Assistant Professor in the B. Thomas Golisano College of Computing and Information Sciences at Rochester Institute of Technology. Prior to that, he visited the National Laboratory of Pattern Recognition (NLPR), Chinese Academy of Science, and the Department of Computer Science and Engineering, University at Buffalo, SUNY. He was a postdoc in the Department of ECE, Northeastern University. Dr. Kong's research interests include computer vision, social media analytics, and machine learning. He is a member of the IEEE.
\vspace{-10mm}
\end{IEEEbiography}

\begin{IEEEbiography}[{\includegraphics[width=1in,height=1.25in,clip,keepaspectratio]{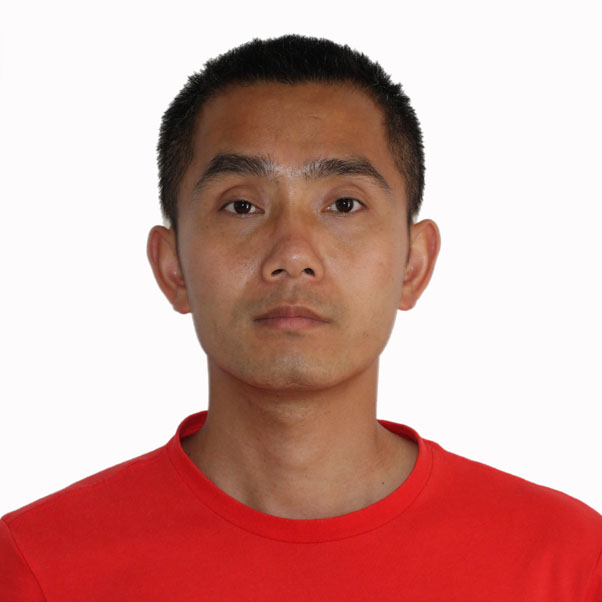}}]{Bineng Zhong} received the B.S., M.S., and Ph.D. degrees in computer science from the Harbin Institute of Technology, Harbin, China, in 2004, 2006, and 2010, respectively. From 2007 to 2008, he was a Research Fellow with the Institute of Automation and Institute of Computing Technology, Chinese Academy of Science. From September 2017 to September 2018, he was a visiting scholar in Northeastern University, Boston, MA, USA. Currently, he is an professor with the School of Computer Science and Technology, Huaqiao University, Xiamen, China. His research interests include pattern recognition, machine learning, and computer vision.
\vspace{-5mm}
\end{IEEEbiography}

\begin{IEEEbiography}[{\includegraphics[width=1in,height=1.25in,clip,keepaspectratio]{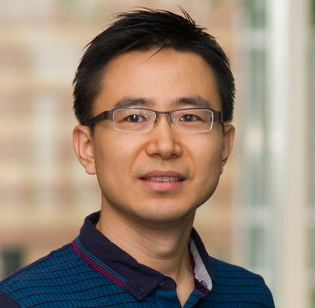}}]{Yun Fu (S'07-M'08-SM'11-F'19)} received the B.Eng. degree in information engineering and the M.Eng. degree in pattern recognition and intelligence systems from Xi’an Jiaotong University, China, respectively, and the M.S. degree in statistics and the Ph.D. degree in electrical and computer engineering from the University of Illinois at Urbana-Champaign, respectively. He is an interdisciplinary faculty member affiliated with College of Engineering and the College of Computer and Information Science at Northeastern University since 2012. His research interests are Machine Learning, Computational Intelligence, Big Data Mining, Computer Vision, Pattern Recognition, and Cyber-Physical Systems. He has extensive publications in leading journals, books/book chapters and international conferences/workshops. He serves as associate editor, chairs, PC member and reviewer of many top journals and international conferences/workshops. He received seven Prestigious Young Investigator Awards from NAE, ONR, ARO, IEEE, INNS, UIUC, Grainger Foundation; nine Best Paper Awards from IEEE, IAPR, SPIE, SIAM; many major Industrial Research Awards from Google, Samsung, and Adobe, etc. He is currently an Associate Editor of the IEEE Transactions on Neural Networks and Leaning Systems (TNNLS). He is fellow of IEEE, IAPR, OSA and SPIE, a Lifetime Distinguished Member of ACM, Lifetime Member of AAAI and Institute of Mathematical Statistics, member of ACM Future of Computing Academy, Global Young Academy, AAAS, INNS and Beckman Graduate Fellow during 2007-2008.
\end{IEEEbiography}
\balance




\end{document}